\documentclass[runningheads]{llncs}
\usepackage[T1]{fontenc}
\usepackage{graphicx}
\usepackage{subcaption}
\usepackage{booktabs} 

\usepackage{amsmath}
\usepackage{amssymb}
\usepackage{mathtools}

\usepackage{multirow}

\usepackage{tikz}

\usepackage{array}
\newcolumntype{M}[1]{>{\centering\arraybackslash}m{#1}}

\usepackage{makecell}

\DeclareCaptionLabelFormat{andtable}{#1~#2  \&  \tablename~\thetable}

\usepackage{hyperref}
\usepackage[depth=3]{bookmark}
\usepackage{color}

\begin{document}

\title{Horizon Activation Mapping for Neural Networks in Time Series Forecasting}
%

\author{Krupakar Hans\inst{1}\thanks{Corresponding author (hansk@nyu.edu)}\orcidID{0000-0002-5072-7633} \and
V A Kandappan\inst{1}\orcidID{0000-0003-4512-7497}}
\authorrunning{K. Hans and V.A. Kandappan}
%
\institute{Department of Computer Science Engineering, \\Shiv Nadar University Chennai, Chennai, India -- 603110 \\
\email{kandappanva@snuchennai.edu.in}}

\maketitle              
\begin{abstract}
Neural networks for time series forecasting have relied on error metrics and architecture-specific interpretability approaches for model selection that don't apply across models of different families. To interpret forecasting models agnostic to the types of layers across state-of-the-art model families, we introduce Horizon Activation Mapping (HAM), a visual interpretability technique inspired by grad-CAM that uses gradient norm averages to study the horizon's subseries where grad-CAM studies attention maps over image data. We introduce causal and anti-causal modes to calculate gradient update norm averages across subseries at every timestep and lines of proportionality signifying uniform distributions of the norm averages. Optimization landscape studies with respect to changes in batch sizes, early stopping, train-val-test splits, architectural choices, univariate forecasting and dropouts are studied with respect to performances and subseries in HAM. Interestingly, batch size based differences in activities seem to indicate potential for existence of an exponential approximation across them per epoch relative to each other. Multivariate forecasting models including MLP-based \emph{CycleNet}, \emph{N-Linear}, \emph{N-HITS}, self attention-based \emph{FEDformer}, \emph{Pyraformer}, SSM-based \emph{SpaceTime} and diffusion-based \emph{Multi-Resolution DDPM} over different horizon sizes trained over the ETTm2 dataset are used for HAM plots in this study. \emph{NHITS}' neural approximation theorem and \emph{SpaceTime}'s exponential autoregressive activities have been attributed to trends in HAM plots over their training, validation and test sets. In general, HAM can be used for granular model selection, validation set choices and comparisons across different neural network model families.
\end{abstract}

\section{Introduction}

Time series forecasting has seen performance improvements because of neural network models similar to those in other modalities in the recent years and these models have been able to forecast over longer horizons. The existence of many families of neural network models that perform well in the forecasting task has motivated arguments for and against models of different families for the task. The arguments address the speed and memory aspects of the forecasting, with performances often remaining similar \cite{zeng2023transformers,zhang2025are}. Foundation models have also been introduced towards the task that are capable of zero-shot generalizability subject to some limitations \cite{cohen2025toto,das2024a}. The model families have yet to find a large enough supervised learning dataset towards comparisons across them and this motivates a framework that can analyze gradient activities in these networks agnostic to the types of layers in them. 

Grad-CAM \cite{selvaraju2017grad} uses gradient update weighted convolutional layer maps in computer vision models in order to find regions attended to over the input image. The use of different types of layers and the spatio-temporal aspects of multivariate forecasting models make analyzing them similarly non-trivial. We introduce Horizon Activation Mapping (HAM), a framework for interpretability of gradient updates in neural network models as a function of subseries by timesteps in the horizon to identify properties of the models with respect to update activity magnitudes. The plots help understand the optimization landscape with respect to the task, enabling interpretable architecture changes as well as compare learning strategies across model variants through the lens of statistical forecasting. Multivariate forecasting based neural network models of different model families have been trained over 60-20-20 splits of the ETTm2 dataset \cite{zhou2021informer}. HAMs have been plotted using gradient norm averages across the entire training dataset using the largest possible batch sizes over an NVIDIA A40 GPU imitating the training and using smaller batch sizes for optimization landscape studies.

Throughout the paper, HAM refers to Horizon Activation Mapping, the overall framework including all the types of plots, and Horizon Activation Map in particular that only includes the causal and anti-causal mode curves. In Section \ref{sec2}, different aspects of the plot are defined and limitations are highlighted. In Section \ref{sec3}, HAMs are studied over horizon size $H=720$ models with respect to changes in dropouts in \emph{NHITS} \cite{challu2023nhits}, batch sizes, epochs after early stopping and dataset splitting in \emph{CycleNet} \cite{lin2024cyclenet} and \emph{NHITS}, the optimization landscape through the training and univariate vs multivariate forecasting in \emph{CycleNet}, and architectural changes in models using \emph{CycleNet}'s cycle length hyperparameter and \emph{N-Linear}'s \cite{zeng2023transformers} normalization. In Section \ref{sec4}, MLPs, self attention, SSM-based and diffusion models of horizon sizes $H=\{96,192,336,720\}$ used in forecasting are studied with respect to different horizon sizes using HAMs across horizon sizes per model and difference plots. While these plots take overall norm averages to interpret models, layer-wise plots also can be used to understand learning dynamics with respect to shorter and longer subseries by layers. The code is available here: \href{https://github.com/hansk0812/Forecasting-Models/tree/lhf}{https://github.com/hansk0812/Forecasting-Models/tree/lhf}. Algorithmic similarities between HAM and Matryoshka Representation Learning (MRL) \cite{kusupati2022matryoshka} validate the use of subseries to represent timesteps in the horizon.

\section{Horizon Activation Mapping (HAM)}
\label{sec2}

Horizon Activation Mapping (HAM) inhibits subseries causally and anti-causally similar to probabilities in grad-CAM \cite{selvaraju2017grad} in order to study regression using subseries-based gradient update magnitudes. Given a horizon of size $H$, the causal mode $c$ includes subseries from $0$ to any timestep $h$ and the anti-causal mode $a$ from any timestep $h$ to $H$. The mask $M$ applied over the loss function before backpropagation is given by  

\begin{equation}
    \begin{split}
        M_c&(\hat{h},H) = \left\{\begin{array}{lr}
                        1, & 1 \leq h \leq \hat{h}, \\
                        0, & \hat{h} < h \leq H
                    \end{array}\right\} \: \forall h \in \mathbb{N} \cap [1,H] \\
        M_a&(\hat{h},H) = \left\{\begin{array}{lr}
                        1, & \hat{h} < h \leq H, \\
                        0, & 1 \leq h \leq \hat{h}
                    \end{array}\right\} \: \forall h \in \mathbb{N} \cap [1,H] \\
        &\mathcal{L}(f_\theta(x^{(g)}), y^{(g)}) = \frac{\sum_{h=1}^H l_h(f_\theta (x^{(g)}_p),y^{(g)}_p)}{H} \\
        \nabla_\theta \mathcal{L}(f_\theta &(x^{(g)}),y^{(g)}, i) = \nabla_\theta \frac{M_m(i,H) \cdot l(f_\theta (x^{(g)}_p),y^{(g)}_p)}{H} 
    \end{split}
    \label{mask}
\end{equation}

where $i \in \mathbb{N} \cap [1,H]$, $m \in \{c,a\}$, $f_\theta$ represents a neural network with learnable parameters $\theta$ and $\mathcal{L}$ denotes its loss function. A line of proportionality connecting $y=0$ and $y=G$ is used in both causal and anti-causal modes corresponding to a uniform distribution of gradient norm averages by the subseries, where $G=\max_{\forall i}(\mu_\theta (\nabla_\theta \mathcal{L}(f_\theta (x^{(g)}),y^{(g)}, i)))$ represents the maximum gradient norm average across all the subseries. For models with auxiliary losses like with \emph{SpaceTime} \cite{zhang2023effectively}, only the regression loss is masked in this study, in which case $M_m$ could also be defined for $0$, with non-zero gradients. The framework also enables studies over other subseries types and masking vs differences in averages.

\subsubsection{Rates of Change}

To magnify the variations in the curves, areas between the curves and the lines of proportionality are calculated using Gauss' formula per timestep such that areas below the lines are negative and changes in slope indicate intersections with them. 

\begin{equation}
    \begin{split}
        A(\hat{h}) = \sum_{\substack{r \in R \\ |R|=a+b}} \left\{\begin{array}{lr}
                        \:\:\:A_P(p_0,p_1,...,p_{s_r}), \forall p_i(y)>p_{l_r}(y), \\
                        -A_P(p_0,p_1,...,p_{s^{'}_r}), \forall p_i(y)<p_{l_r}(y)
                    \end{array}\right\}, \\ \forall p_i(x) \leq \hat{h}
    \end{split}
    \label{areas}
\end{equation}
where there are $a$ regions above and $b$ regions below the lines of proportionality $l$,  $p_i \in \mathbf{R}^2$ represents points in the curves, $l_r$ represents points of intersections of regions $r$ on the right, $A_P$ is calculated using the shoelace formula \cite{boland2000polygon,contreras1998cutting} and $A(\hat{h})$ applies to both causal and anti-causal mode curves. Where this definition helps with amplifying the differences, it accounts for transitions from below the proportionality line to above it as a continuum and this creates magnitude lags in timesteps between positive and negative curves, making comparisons across models a bit tedious.

\subsubsection{Gradient Equivariant Point}

The causal and anti-causal mode curves intersect at the gradient equivariant point $h$ in the horizon of length $H$ where the first $h$ timesteps' gradient norm average is the same as the last $H-h$ timesteps'. 

\subsubsection{Interpolated Area Plots}

To compare the gradient activities of single models of different horizon sizes, the maximum area per model scales the plots to the $[0,1]$ range. 

\subsubsection{Difference Plots}

For every timestep in the horizon, the difference between the gradient norm averages in the first and the subsequent subseries scaled to $[-1,1]$ helps compare across model families. The gradient equivariant points are highlighted in the difference plots as points along the $y=-1$ line. Given the gradient norm average function over the weights $g_\theta: \mathbf{R}^2 \rightarrow \mathbf{R}$ from timesteps $t_1$ to $t_2$, the difference function is given by
\begin{equation}
    d(t) = \frac{g_\theta(0 \rightarrow t) - g_\theta(t \rightarrow H)}{\max_{\forall t} g_\theta(0 \rightarrow t) - g_\theta(t \rightarrow H)}, t \in \mathbb{W} \cap [0,H]
    \label{diff_eq}
\end{equation}





\section{Optimization Landscape Studies}
\label{sec3}

\subsection{Dropouts}

\begin{figure}[!htbp]
    \centering
    \includegraphics[width=0.495\linewidth,height=2.1675cm]{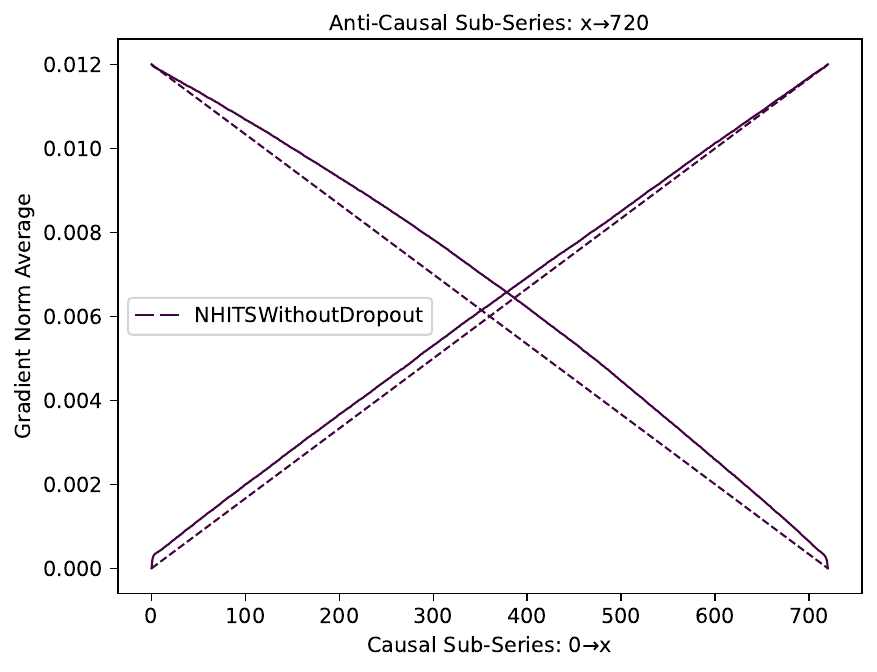} \hfill 
    \includegraphics[width=0.495\linewidth,height=2.1675cm]{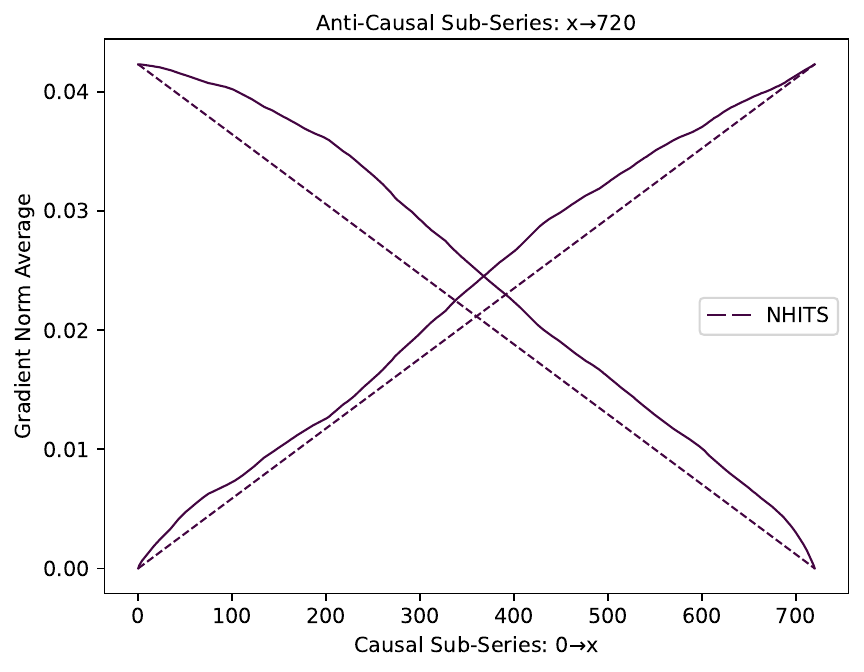} \hfill \\
    \subfloat[\centering NHITS Without Dropout]{\includegraphics[width=0.495\linewidth,height=2.1675cm]{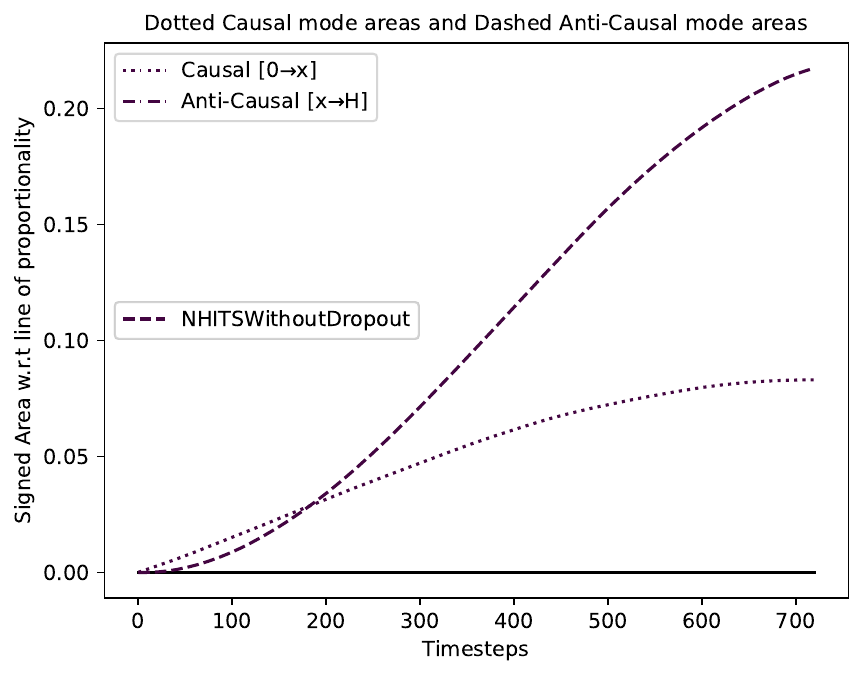}} \hfill
    \subfloat[\centering NHITS]{\includegraphics[width=0.495\linewidth,height=2.1675cm]{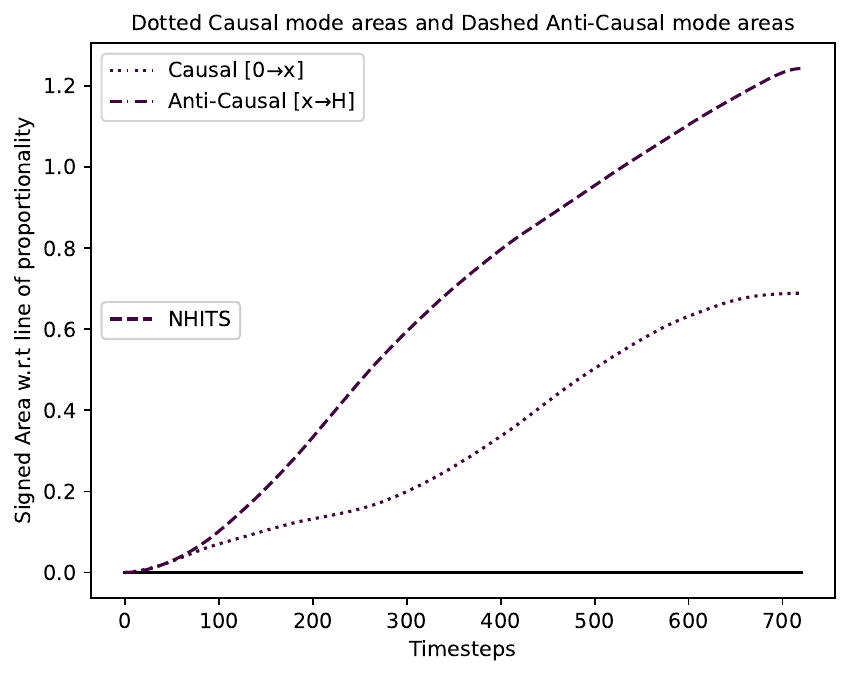}} \hfill
    \caption{$H=720$ \emph{NHITS} gradient activities with and without a dropout of 0.2 chosen based on performance.}
    \label{dropouts}
\end{figure}

In general, dropouts helped more with multivariate forecasting models than univariate ones. Dropouts do to the feature space what HAM does to the output data space. From studies on \emph{NHITS} in Figure \ref{dropouts}, it is evident that the addition of dropout increases the overall gradient magnitudes by more than 3 times, and gradient magnitudes in the longer regions of the horizon.

\subsection{Batch Size Differences}

\begin{figure}[!htbp]
    \centering 
    \subfloat[\centering Error metrics]{\begin{minipage}{0.21\linewidth}
    \vskip-55pt
    \begin{tabular}{ccc} \toprule 
    Batch & MSE & MAE \\
    \midrule 
    4000 & 0.225 & 0.340 \\
    2000 & 0.227 & 0.344 \\
    500 & 0.230 & 0.346 \\
     \bottomrule
    \end{tabular}
    \end{minipage}} \hfill 
    \subfloat[\centering HAM]{\includegraphics[width=0.39\linewidth,height=2.1675cm]{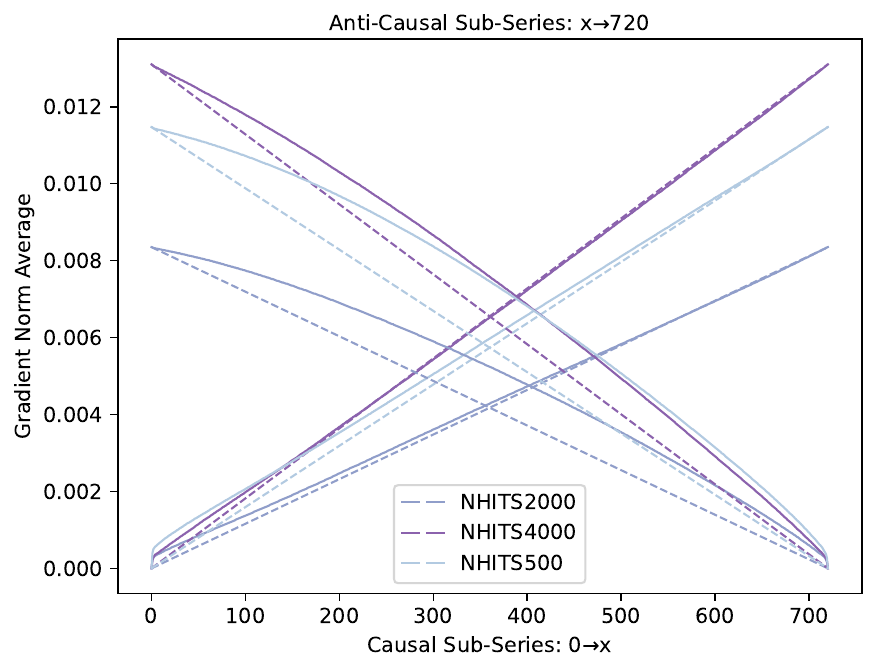}} \hfill
    \subfloat[\centering Area curves]{\includegraphics[width=0.39\linewidth,height=2.1675cm]{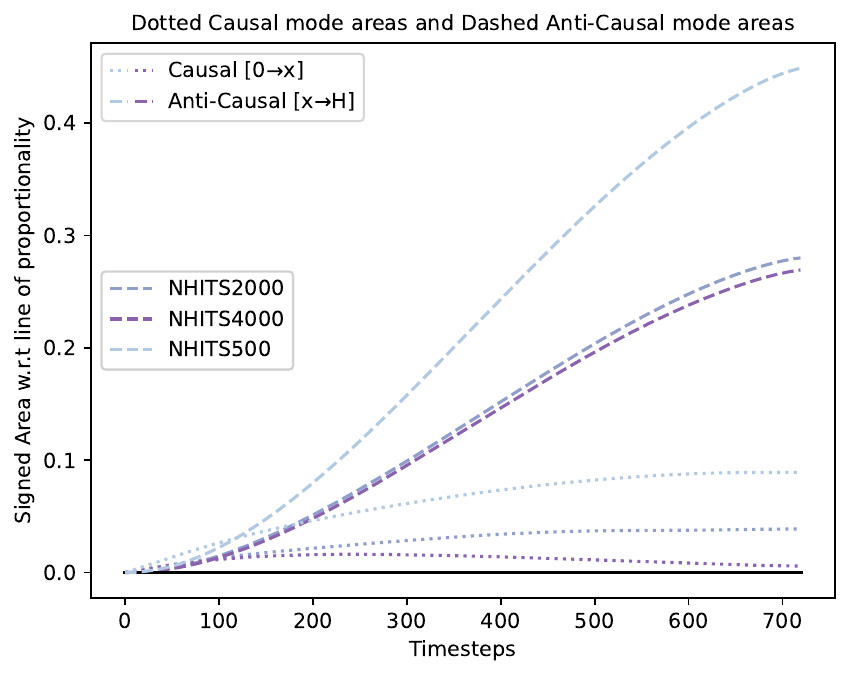}} \hfill
    \captionlistentry[table]{HAMs of $H=720$ \emph{NHITS} models with different batch sizes show their variations by norm average magnitudes. The plots get darker corresponding to increases in batch sizes.}
    \captionsetup{labelformat=andtable}
    \caption{HAMs of $H=720$ \emph{NHITS} models with different batch sizes. The plots get darker corresponding to increases in batch sizes.}
    \label{bs_fig}
\end{figure}

\begin{figure}[!htbp]
    \centering
    \includegraphics[width=0.495\linewidth,height=2.1675cm]{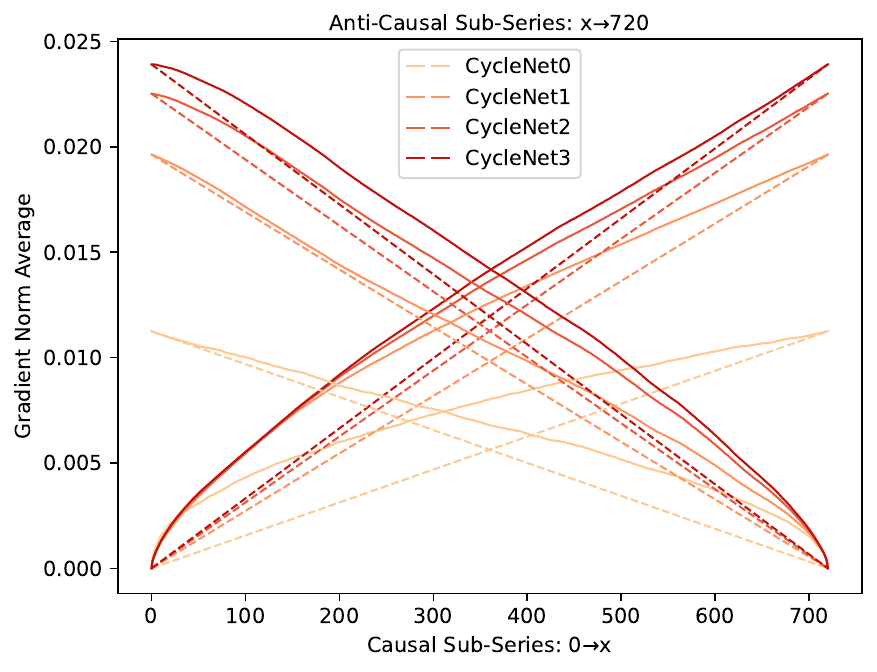} \hfill 
    \includegraphics[width=0.495\linewidth,height=2.1675cm]{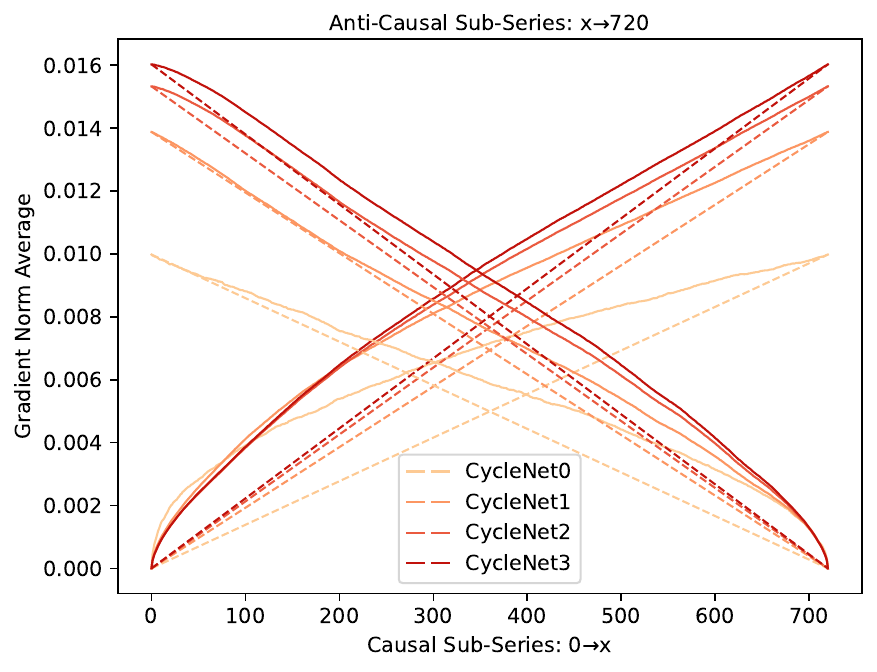} \hfill \\ 
    \subfloat[\centering Batch Size 50]{\includegraphics[width=0.495\linewidth,height=2.1675cm]{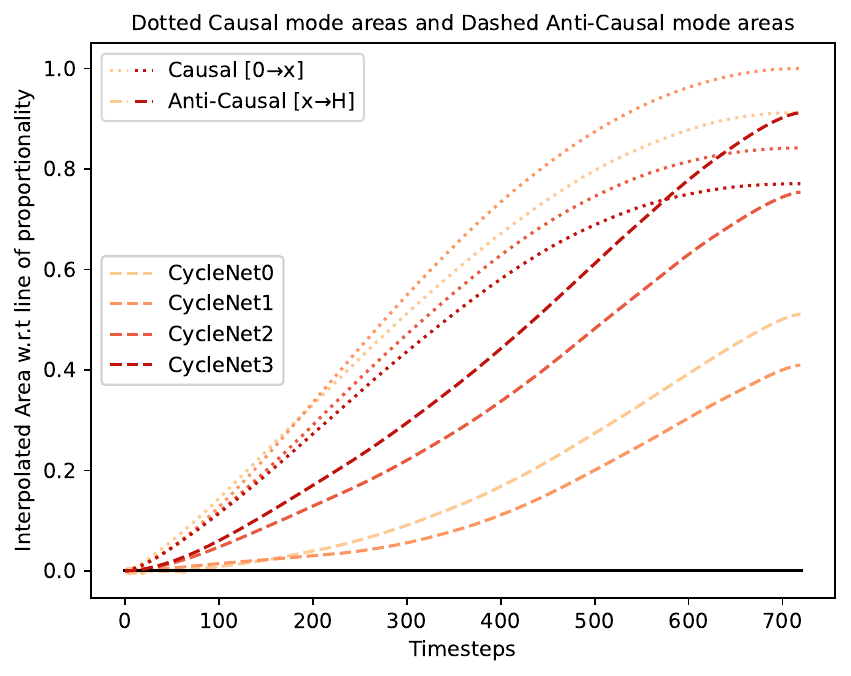}} \hfill
    \subfloat[\centering Batch Size 100]{\includegraphics[width=0.495\linewidth,height=2.1675cm]{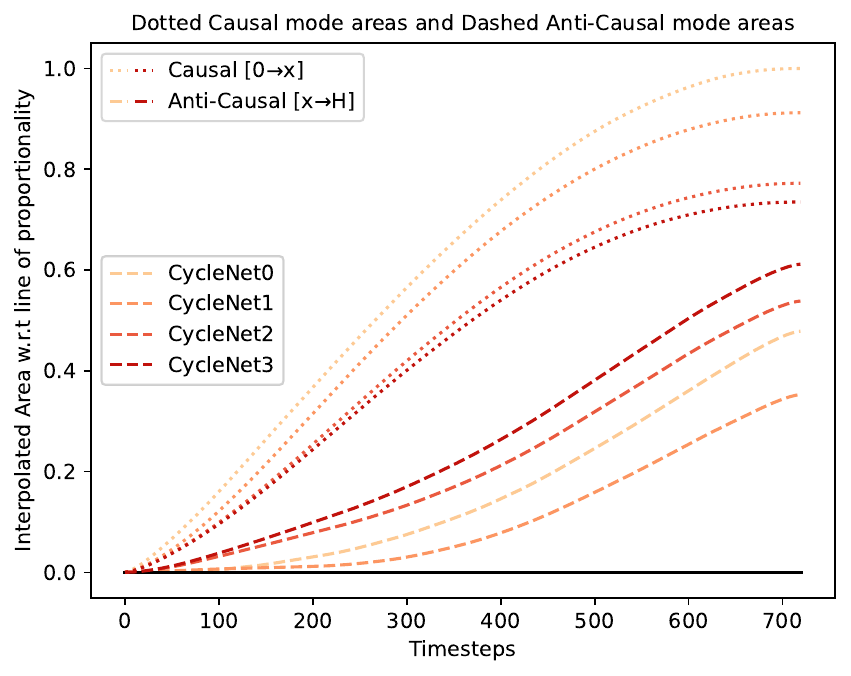}} \hfill
    \caption{$H=720$ \emph{CycleNet} models with different batch sizes show norm averages increasing over number of epochs. The numbers next to \emph{CycleNet} correspond to epochs with 0 indicating the randomly initialized model's. The plots are darker corresponding to increases in epochs.}
    \label{bs_cyclenet_fig}
\end{figure}

\begin{figure}[!htbp]
    \centering
    \subfloat[\centering NHITS]{\includegraphics[width=0.327\linewidth,height=2.1675cm]{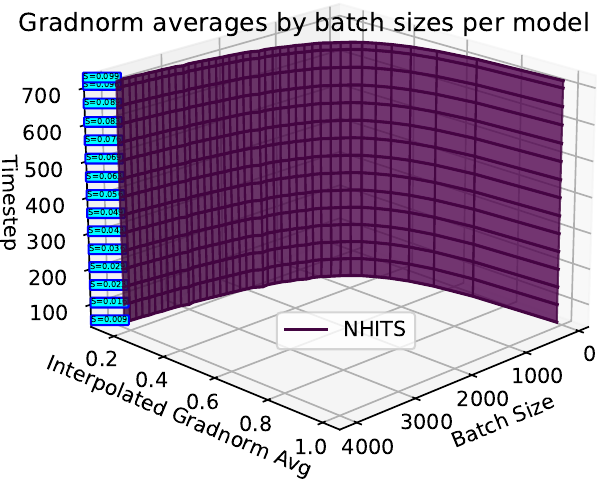}} \hfill 
    \subfloat[\centering N-Linear]{\includegraphics[width=0.327\linewidth,height=2.1675cm]{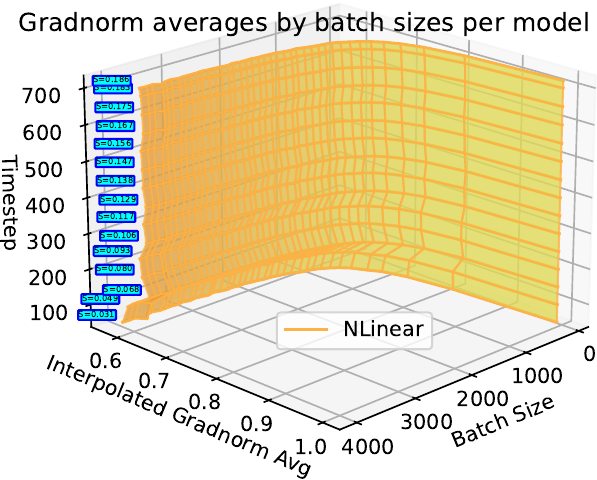}} \hfill 
    \subfloat[\centering CycleNet]{\includegraphics[width=0.327\linewidth,height=2.1675cm]{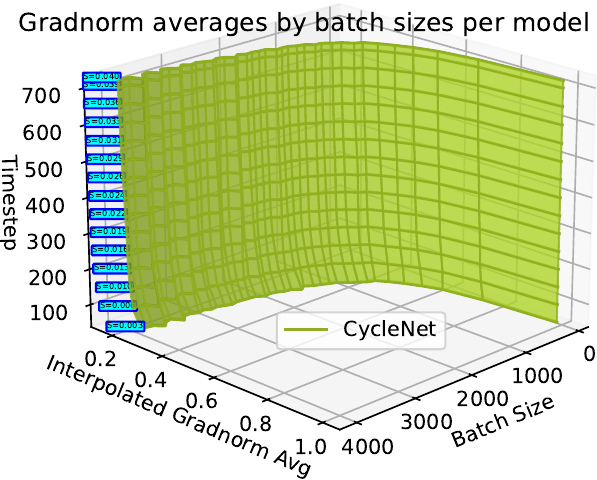}}
    \caption{Although different in HAM magnitudes when batch sizes change over more than one epoch between models, over converged \emph{N-Linear}, \emph{NHITS} and \emph{CycleNet} models, gradient norm averages interpolated by batch size show a polynomially saturating trend as batch sizes increase. This suggests that HAM plots could find polynomial approximations across batch sizes one epoch at a time.}
    \label{batchsizes}
\end{figure}

From Figure \ref{bs_fig}, in terms of magnitude ranges, \emph{NHITS}' activities are in increasing orders of batch sizes 2000, 500, 4000, with errors in descending order, and alongwith both curves saturating closer to lines of proportionality and anti-causal curves higher, highlight activity magnitude increases necessary in longer regions of the horizon as batch sizes increase. \emph{NHITS}' gradient equivariant points decrease in timesteps as the batch sizes increase, but increase in magnitudes. 

Studies can highlight architectural aspects of learning when performances are similar across batch sizes. From Figure \ref{bs_cyclenet_fig}, \emph{CycleNet}'s third epoch increases more anti-causally when batch size is 50 and decreases more causally when batch size is 100, indicating the subtle differences in performance. Increases in range across epochs are greater when its batch size is smaller and vice versa. Its converged model's relative increases are causal when batch size is 100 and anti-causal when batch size is 50. Batch size shows a consistent decrease in causal and anti-causal mode magnitudes in both \emph{CycleNet} and \emph{NHITS}.

Even though there is variability in magnitudes across batch sizes in Figures \ref{bs_fig} and \ref{bs_cyclenet_fig}, Figure \ref{batchsizes} shows a surprising polynomial curve across batch sizes in converged models over a single epoch, motivating future work.

\subsection{Early Stopping}



\begin{figure}[!htbp] 
    \subfloat[\centering HAM]{\includegraphics[width=0.327\linewidth,height=2.1675cm]{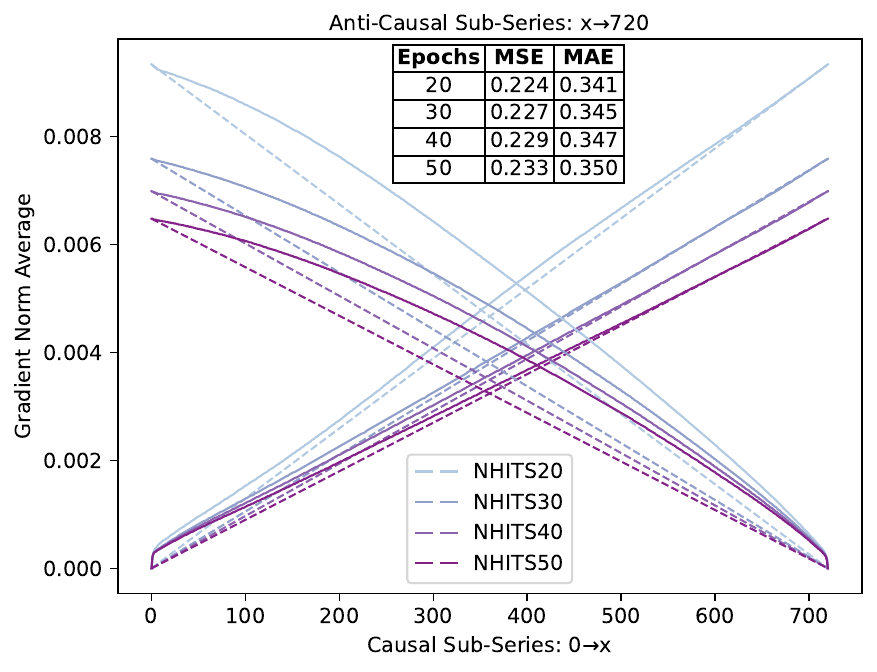}} \hfill 
    \subfloat[\centering Area curves]{\includegraphics[width=0.327\linewidth,height=2.1675cm]{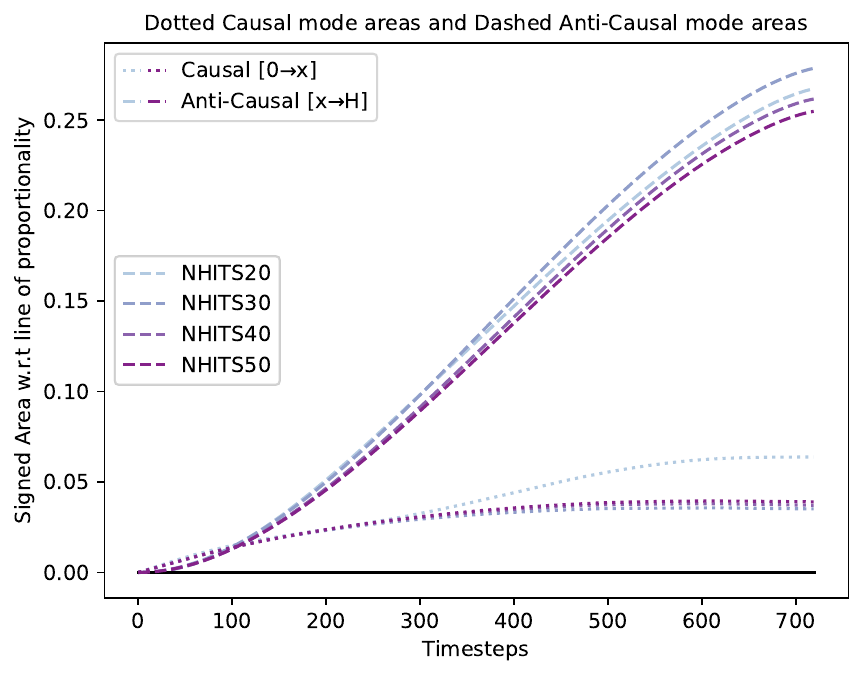}} \hfill
    \subfloat[\centering Difference plot]{\includegraphics[width=0.327\linewidth,height=2.1675cm]{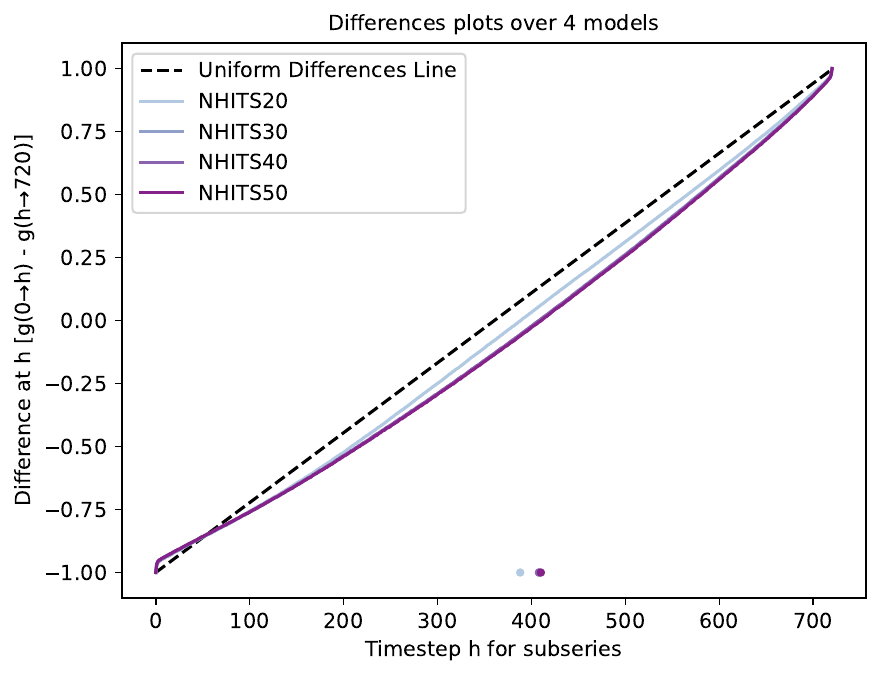}}    
    \captionlistentry[table]{$H=720$ \emph{NHITS} models after early stopping show the reduction in norm average magnitudes proportional to how far the plots are from the early stopped model. The plots get darker corresponding to increases in epoch counts.}
    \captionsetup{labelformat=andtable}
    \caption{$H=720$ \emph{NHITS} models after early stopping show the reduction in norm average magnitudes away from the local minima. The plots get darker corresponding to increases in epoch counts.}
    \label{es_fig}
\end{figure}

Figure \ref{es_fig} shows the gradient norm averages reducing in magnitude as the number of epochs after early stopping increases, indicating that increase in errors correspond to magnitude reductions in early stopping. From the HAM and area curves, the causal mode curves reduce in magnitude after early stopping, first steeply and then gradually, with the magnitude reductions corresponding to being in a local minima. Similarly, all the curves in the difference plot are below the uniform line and that of the early stopped model is closest to it throughout the horizon. The gradient equivariant point of the early stopped model is closer to $\frac{H}{2}$ than the other models' suggesting that activities in longer regions of the horizon don't improve the performance.

\subsection{Dataset Splits}

\begin{figure}[!htbp]
    \centering
    \subfloat[\centering \emph{CycleNet} Random Model]{\includegraphics[width=0.495\linewidth,height=2.1675cm]{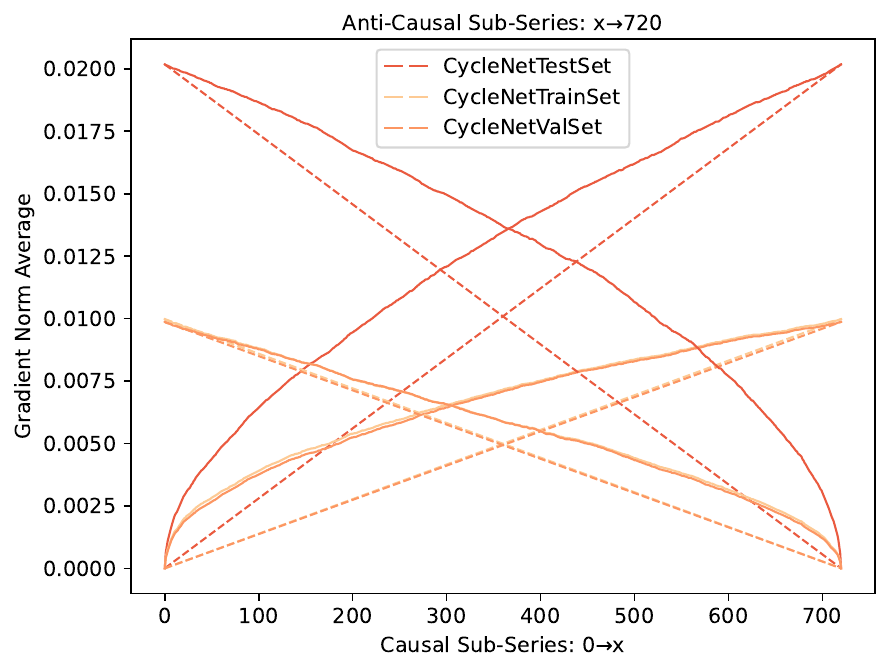}} \hfill
    \subfloat[\centering \emph{CycleNet} Converged Model]{\includegraphics[width=0.495\linewidth,height=2.1675cm]{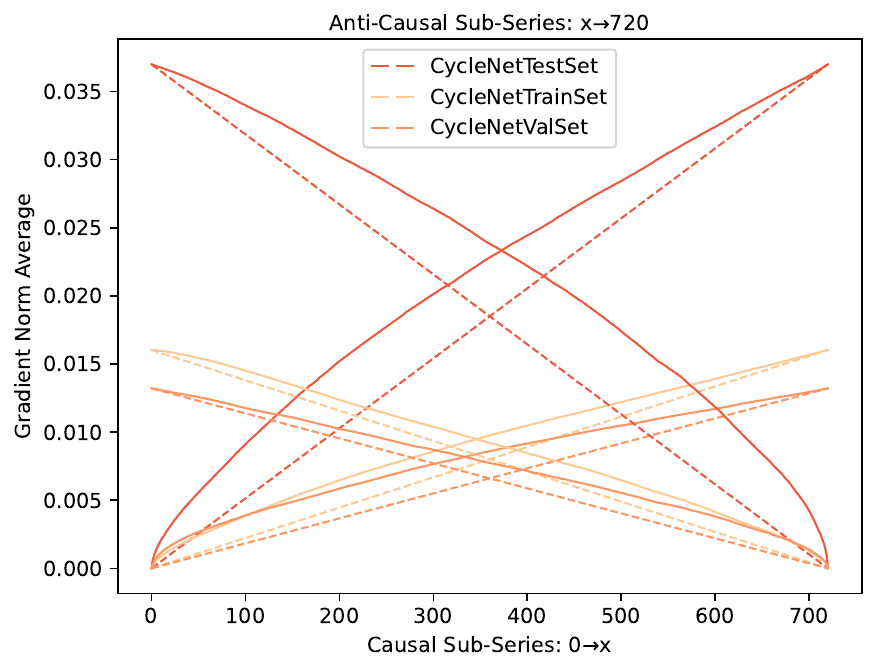}} \\ 
    \subfloat[\centering \emph{NHITS} Random Model]{\includegraphics[width=0.495\linewidth,height=2.1675cm]{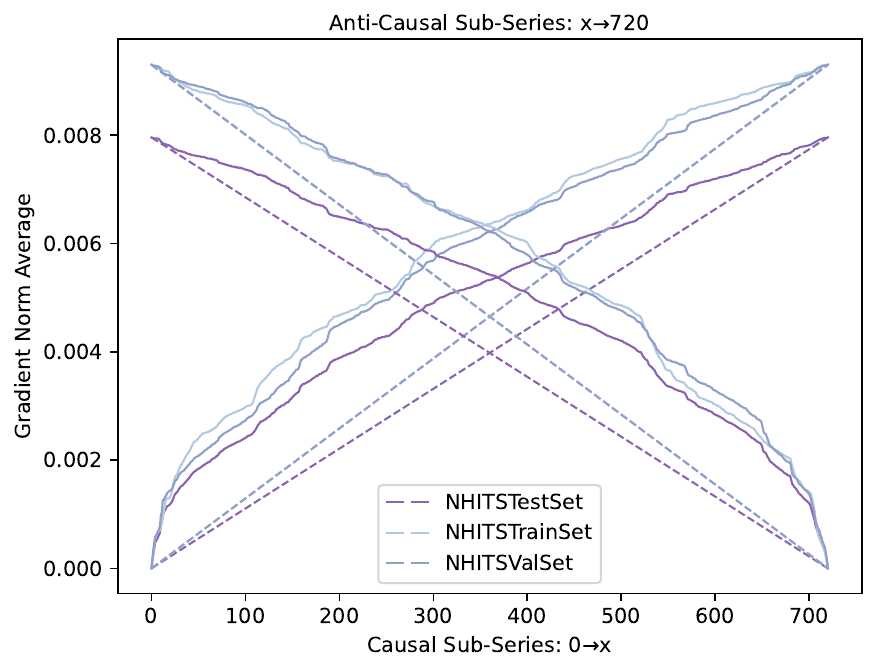}} \hfill
    \subfloat[\centering \emph{NHITS} Converged Model]{\includegraphics[width=0.495\linewidth,height=2.1675cm]{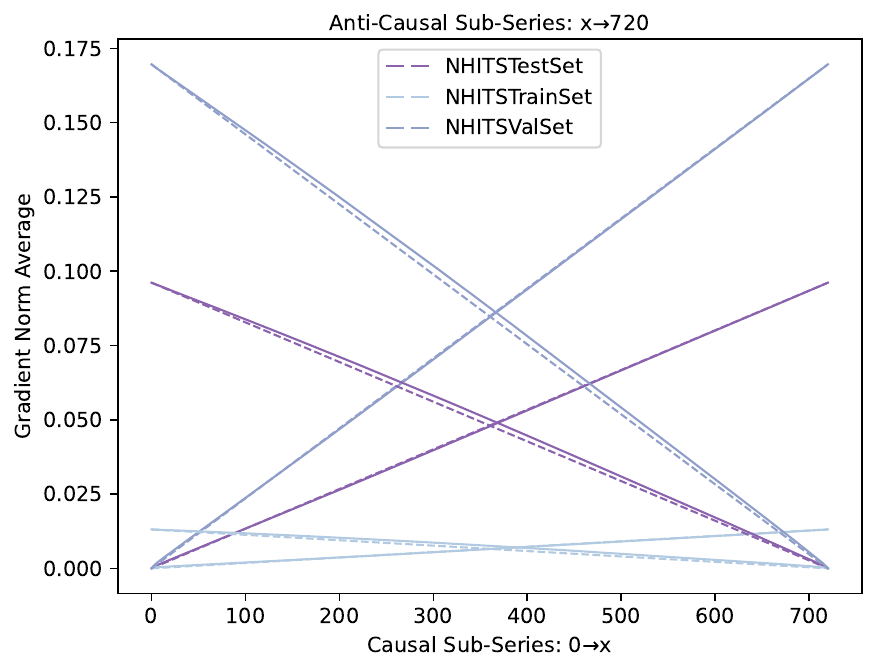}} \hfill
    \caption{HAMs across different dataset splits between randomly initialized and converged models of \emph{NHITS} and \emph{CycleNet}. The plots get darker corresponding to increases in epoch counts.}
    \label{splits}
\end{figure}

While the validation and test sets are not used for gradient updates during the training, variations in HAMs can be used to choose validation sets based on the relative variabilities across timesteps (Figure \ref{splits}). Training pushes the curves of both \emph{CycleNet} and \emph{NHITS} closer to the lines of proportionality and shows increases in magnitudes overall with decaying learning rates. \emph{CycleNet}'s training set's magnitudes are higher than the validation set's magnitudes in its converged model. \emph{NHITS}' update magnitudes seem to indicate out-of-distribution aspects of the validation set more than \emph{CycleNet}'s do. It is noteworthy that \emph{NHITS}' curves over the randomly initialized model show variations with respect to the pooling and interpolation operations in the layers that don't reflect in the converged model's activities.

\subsection{Univariate and Multivariate Variants}

\begin{figure}[!htbp]
    \centering
    \subfloat[\centering Multivariate \emph{CycleNet}]{\includegraphics[width=0.327\linewidth,height=2.1675cm]{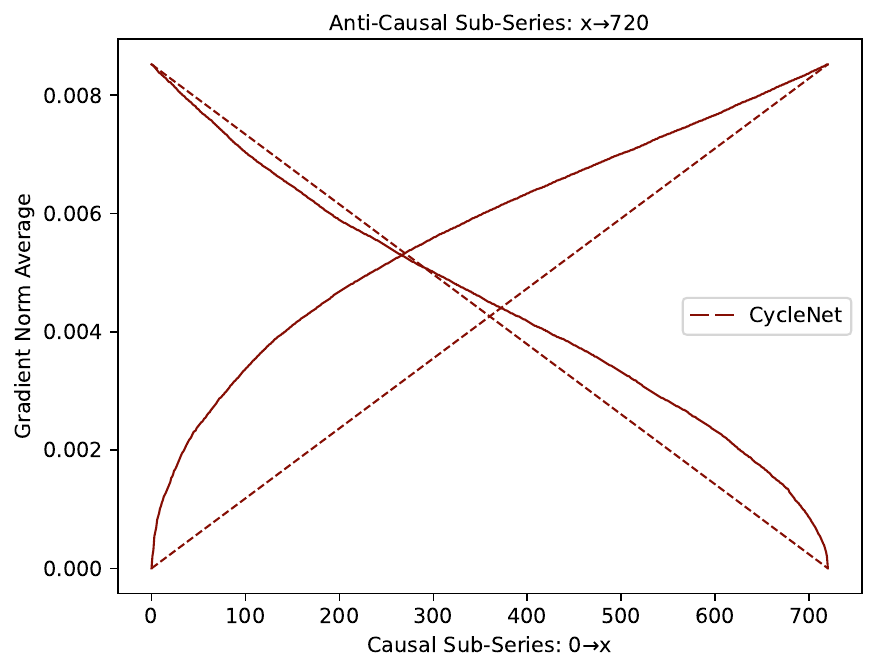}} \hfill 
    \subfloat[\centering Univariate \emph{CycleNet}]{\includegraphics[width=0.327\linewidth,height=2.1675cm]{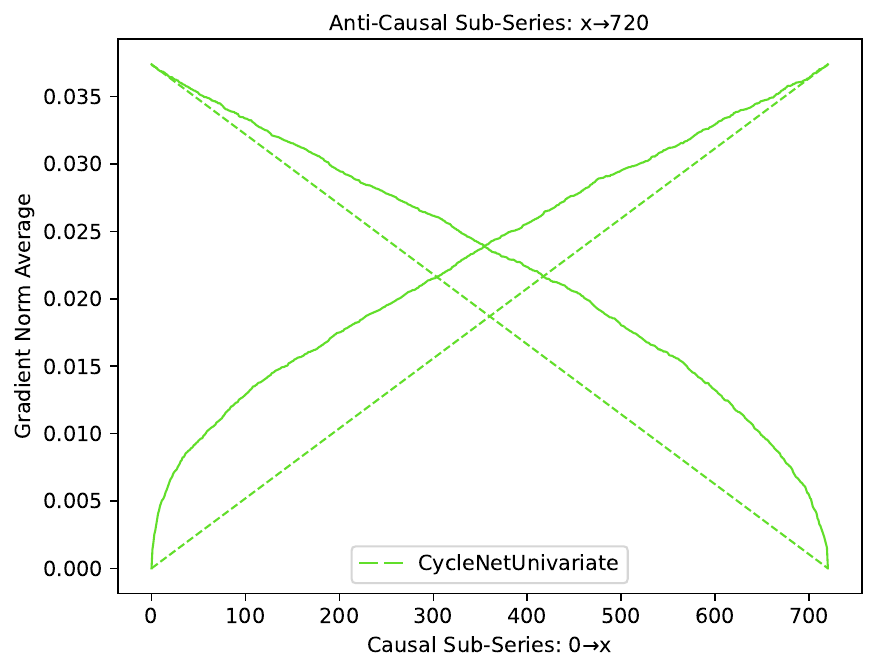}} \hfill
    \subfloat[\centering Interpolated Area Plot]{\includegraphics[width=0.327\linewidth,height=2.1675cm]{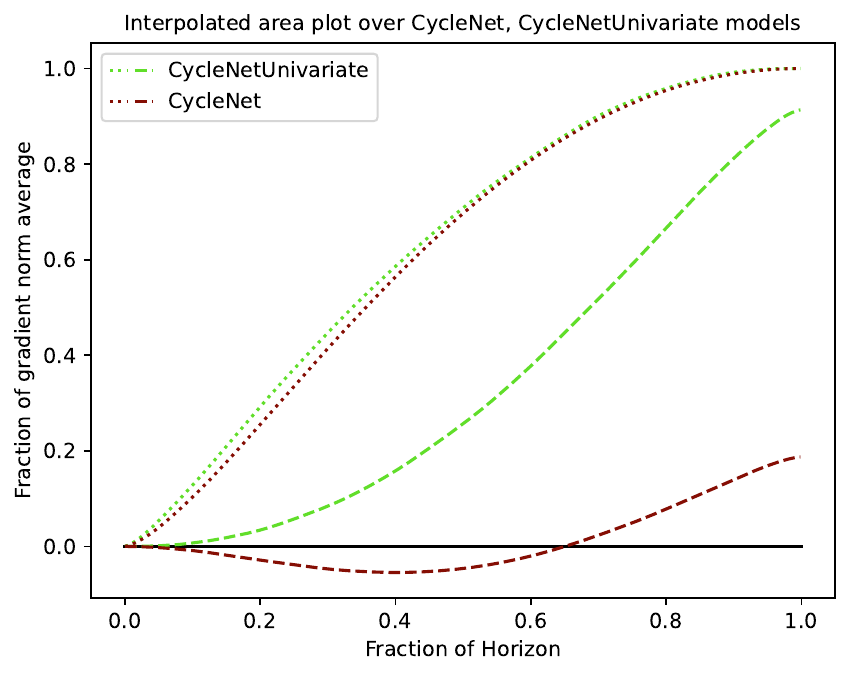}} \hfill 
    \caption{\emph{CycleNet}'s multivariate and univariate High UseFul Load series' HAMs highlight the difficulties involved in the spatial aspects of long horizon forecasting.}
    \label{variates}
\end{figure}

The range differences in Figure \ref{variates} indicate the difficulties in the spatial aspects of learning in multivariate data. Even though the multivariate causal mode curve is farther away from being proportional relative to the univariate curve, the interpolated area plot finds them closer with slight increases that don't reflect beyond shorter horizon subseries. The removal of the initial timesteps from multivariate subseries pushes its anti-causal curve below the line of proportionality where it doesn't in the univariate case and the relative magnitudes are lesser overall relatively as well, as in the interpolated area plot. 

\subsection{Architecture Changes}

\begin{figure}[t]
    \centering
    \subfloat[\emph{N-Linear}]{\includegraphics[width=0.327\linewidth,height=2.1675cm]{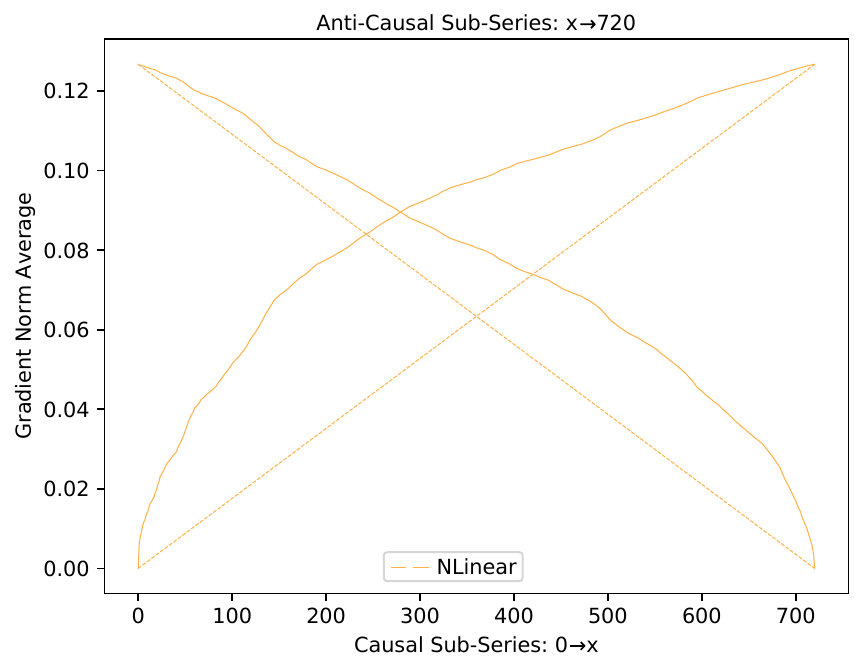}} \hfill
    \subfloat[\centering \emph{N-Linear} without normalization]{
    \includegraphics[width=0.327\linewidth,height=2.1675cm]{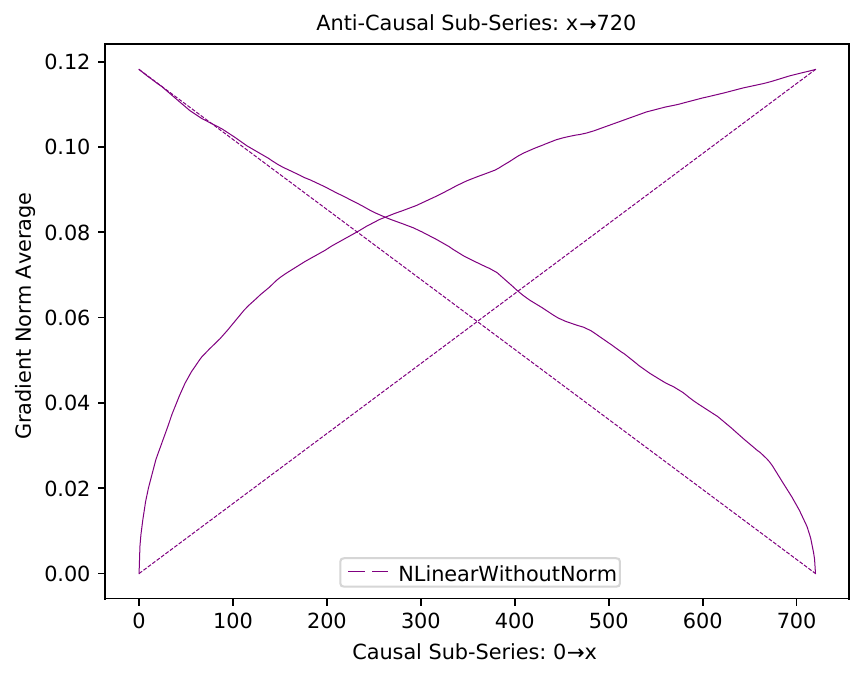}} \hfill
    \subfloat[\centering Interpolated Gradient Norms Plot]{
    \includegraphics[width=0.327\linewidth,height=2.1675cm]{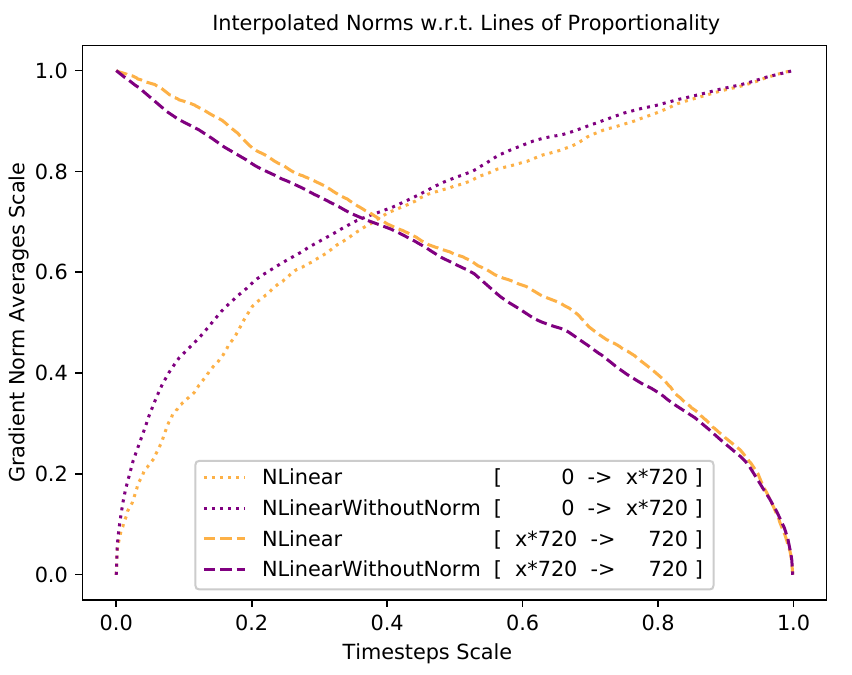}} \\
    \subfloat[\centering \emph{CycleNet}'s Cycle Queue's HAM]{
    \includegraphics[width=0.49\linewidth,height=2.1675cm]{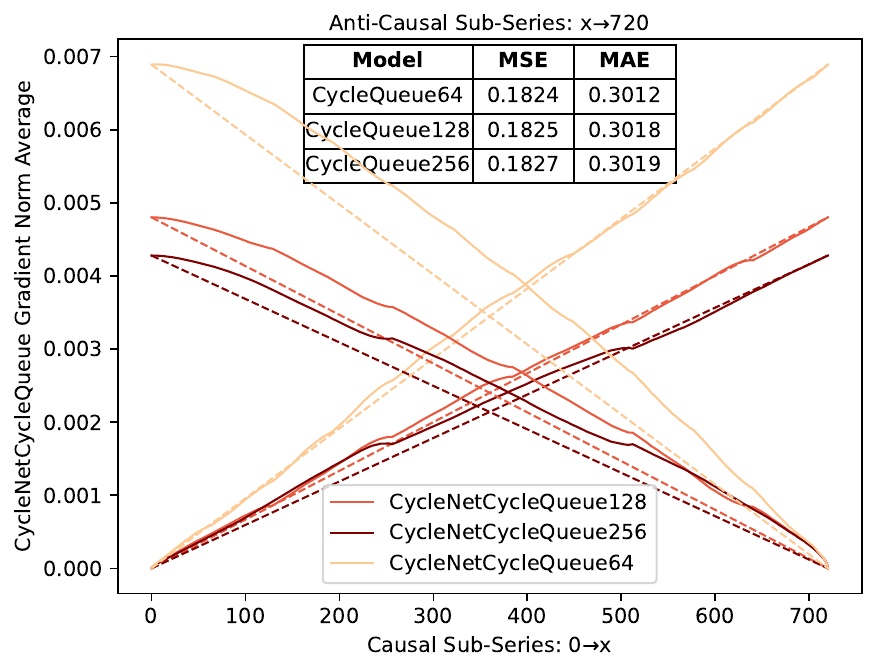}} \hfill
    \subfloat[\centering Cycle Queue's Difference Plot]{
    \includegraphics[width=0.49\linewidth,height=2.1675cm]{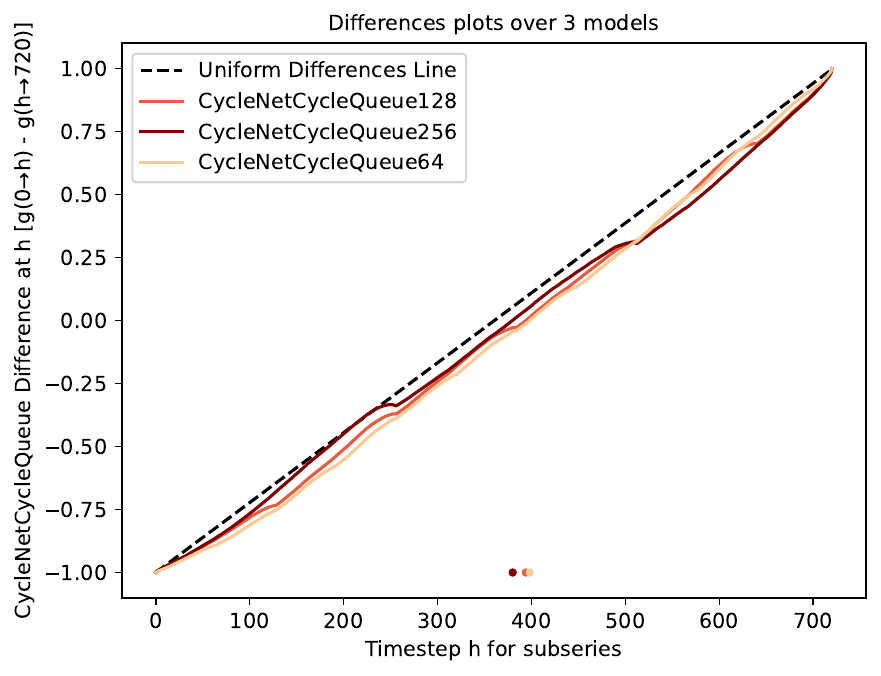}}
    \caption{HAMs and Difference Plots for $H=720$ \emph{N-Linear} with and without its normalization and $H=720$ \emph{CycleNet}'s Cycle Queue of $|Q| \in \{64, 128, 256\}$.}
    \label{nlinear}
\end{figure}


\emph{N-Linear}'s normalization strategy using the last timestep-based difference has reduced its errors by 16\% in our ablation experiments and \emph{CycleNet}'s cycle queue's length decides ordering of its values by dataset-level statistics and we study their HAMs in Figure \ref{nlinear}. In \emph{N-Linear}, subtracting the last timestep's value decreased magnitudes causally over the first 23\% of the horizon and increased it after that. Anti-causal magnitude increases show correspondence to the causal decrease and also happen later in the horizon, deviating from linear trends without normalization. \emph{CycleNet}'s cycle queue's norms have been plotted in HAMs instead of all the layers' to prevent too much averaging. Both the HAM and difference plots show spikes corresponding to lengths of the cycle and intersect with lines of proportionality very close to them when they do. The smallest cycle length corresponds to the best performing model and to its difference plot shows larger magnitudes interplaying between subseries length and longer regions.

\section{Model Families}
\label{sec4}

HAMs have been plotted over multivariate forecasting over ETTm2 using MLP-based \emph{NHITS}, self attention-based \emph{FEDformer} \cite{zhou2022fedformer} and \emph{Pyraformer} \cite{liu2021pyraformer}, SSM-based \emph{SpaceTime} \cite{zhang2023effectively} and diffusion-based \emph{Multi-Resolution DDPM} \cite{shen2024multi}. 

\subsection{Horizon Size Comparisons}


From Table \ref{table8}, in terms of magnitudes, the areas of anti-causal mode gradient norm averages always converge to 1 where all those of the causal mode norm averages don't in \emph{NHITS}, and vice versa in \emph{Pyraformer}, and given the non-autoregressive anti-causal increase trends in them across different subseries against causal ones increasing in horizon sizes, supports the infinite horizon and linear combination assumptions in the neural basis approximation theorem for \emph{NHITS}. The increases in gradients causally over longer regions in $H=720$ \emph{NHITS} and $H=96$ \emph{Pyraformer} models highlight the hierarchical aspects of the models. \emph{Pyraformer}'s curves increase in magnitudes causally and decrease anti-causally as horizon sizes increase over all its models whereas it only happens in \emph{FEDformer}'s $H=\{336,720\}$ models, highlighting its frequency domain based improvements being better performing than pyramidal improvements to self-attention. Further, \emph{FEDformer}'s fourier and wavelet transforms linearize its activities as horizon sizes increase. 

\setlength{\tabcolsep}{0pt}
\begin{table}[t]
    \centering
    \begin{tabular}{cM{2.23cm}M{2.23cm}M{2.23cm}M{2.23cm}M{2.23cm}}
        \toprule
        \textbf{Model} & \textbf{H=96} & \textbf{H=192} & \textbf{H=336} & \textbf{H=720} & \textbf{Areas} \\ 
        \midrule        
        NHITS & \begin{tikzpicture} \node[inner sep=0pt] {
            \includegraphics[width=2.23cm,height=2.1675cm]{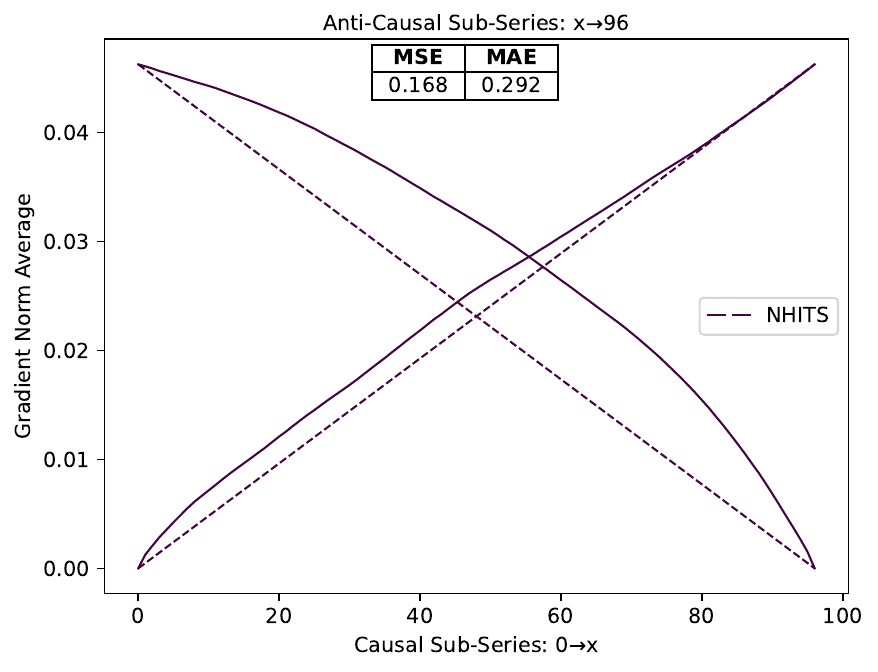}
        };\end{tikzpicture} & \begin{tikzpicture} \node[inner sep=0pt] {
            \includegraphics[width=2.23cm,height=2.1675cm]{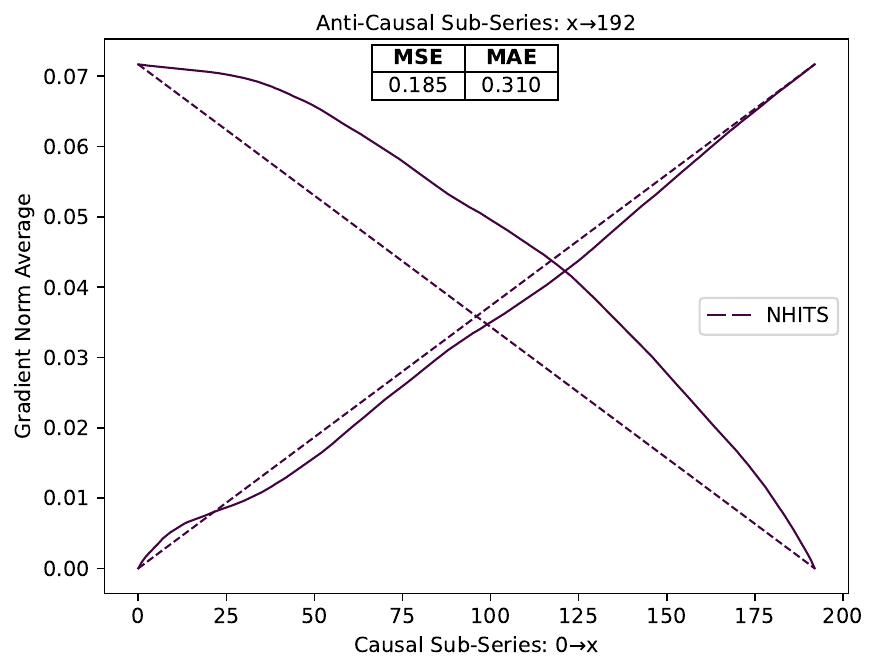}
        };\end{tikzpicture} & \begin{tikzpicture} \node[inner sep=0pt] {
            \includegraphics[width=2.23cm,height=2.1675cm]{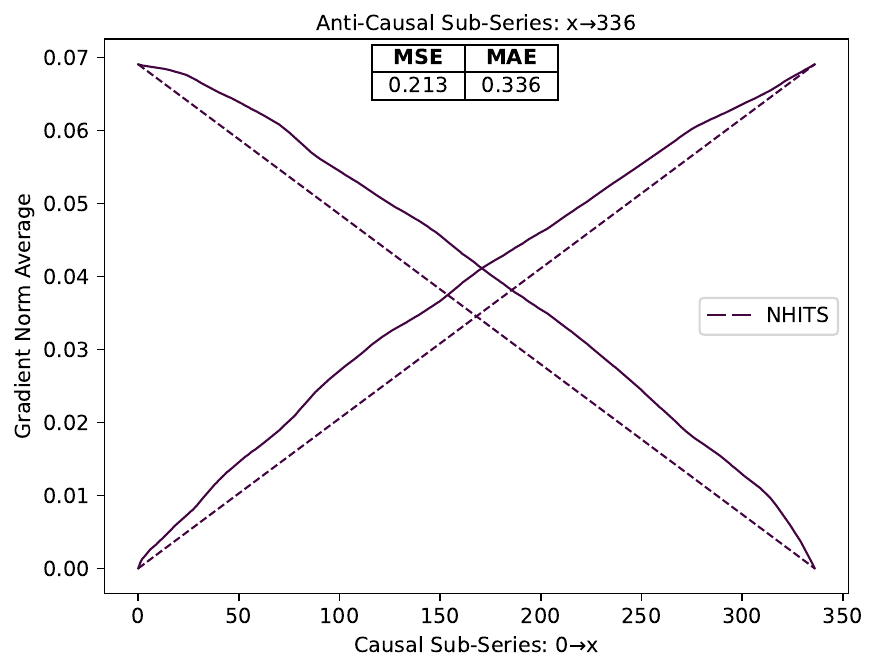}
        };\end{tikzpicture} & \begin{tikzpicture} \node[inner sep=0pt] {
            \includegraphics[width=2.23cm,height=2.1675cm]{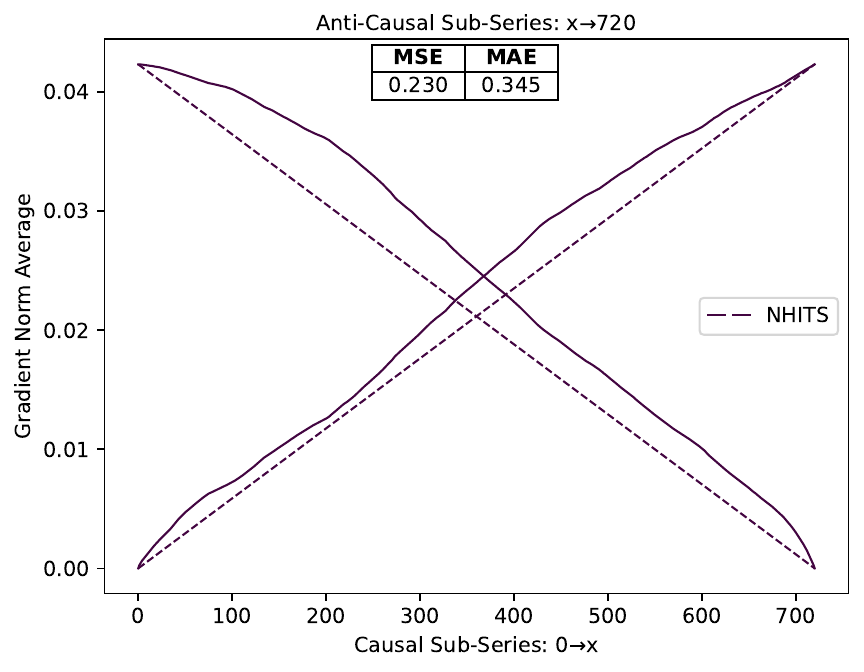}  
        };\end{tikzpicture} & \begin{tikzpicture} \node[inner sep=0pt] {
            \includegraphics[width=2.23cm,height=2.1675cm]{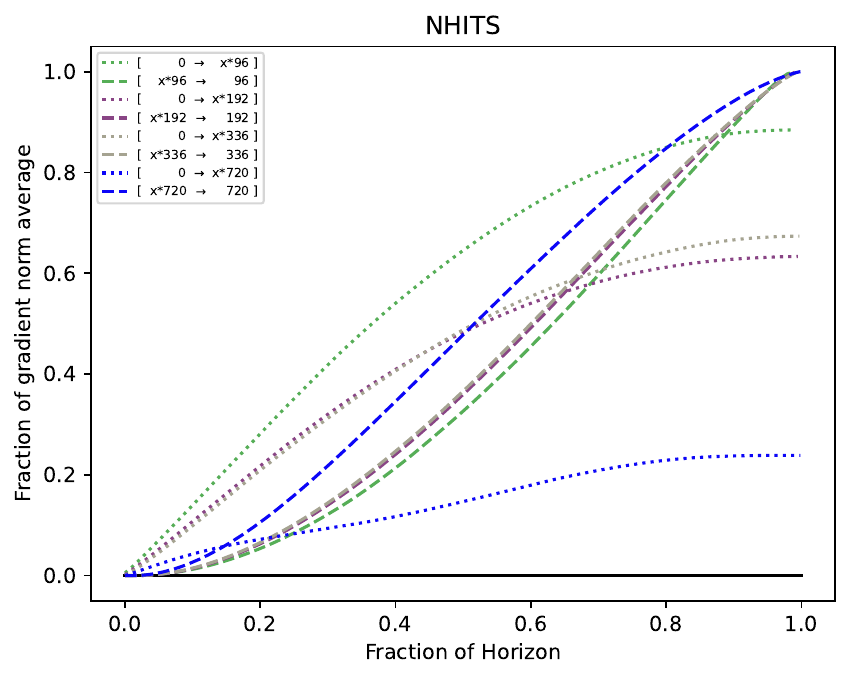}  
        };\end{tikzpicture} \\
        \makecell{FED-\\former} & \begin{tikzpicture} \node[inner sep=0pt] {
            \includegraphics[width=2.23cm,height=2.1675cm]{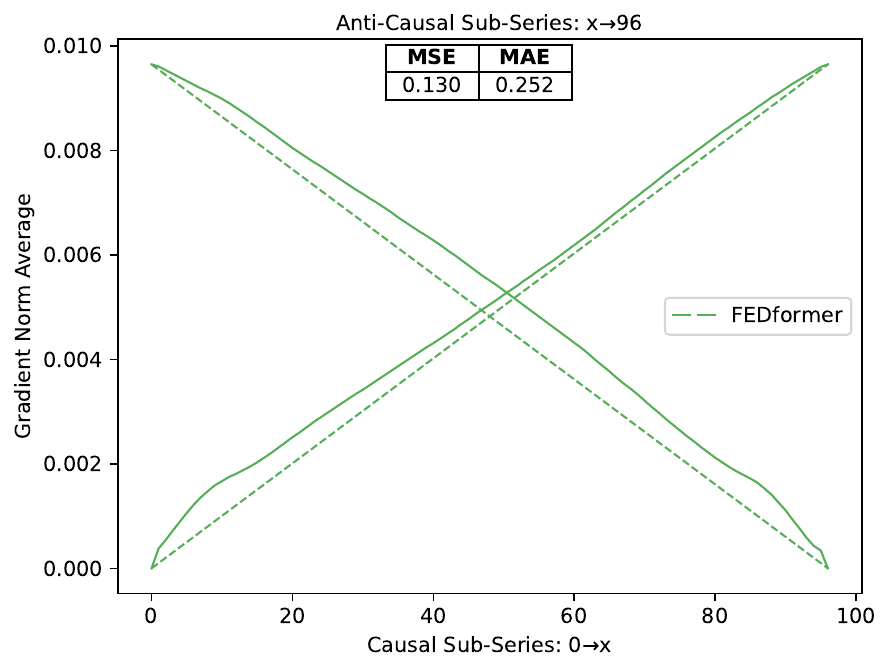}
        };\end{tikzpicture} & \begin{tikzpicture} \node[inner sep=0pt] {
            \includegraphics[width=2.23cm,height=2.1675cm]{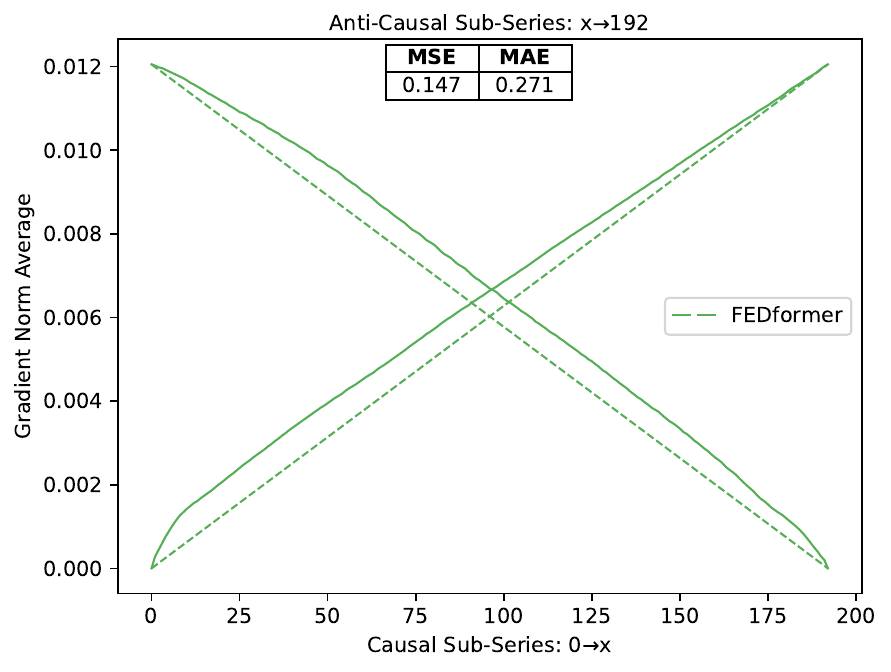}
        };\end{tikzpicture} & \begin{tikzpicture} \node[inner sep=0pt] {
            \includegraphics[width=2.23cm,height=2.1675cm]{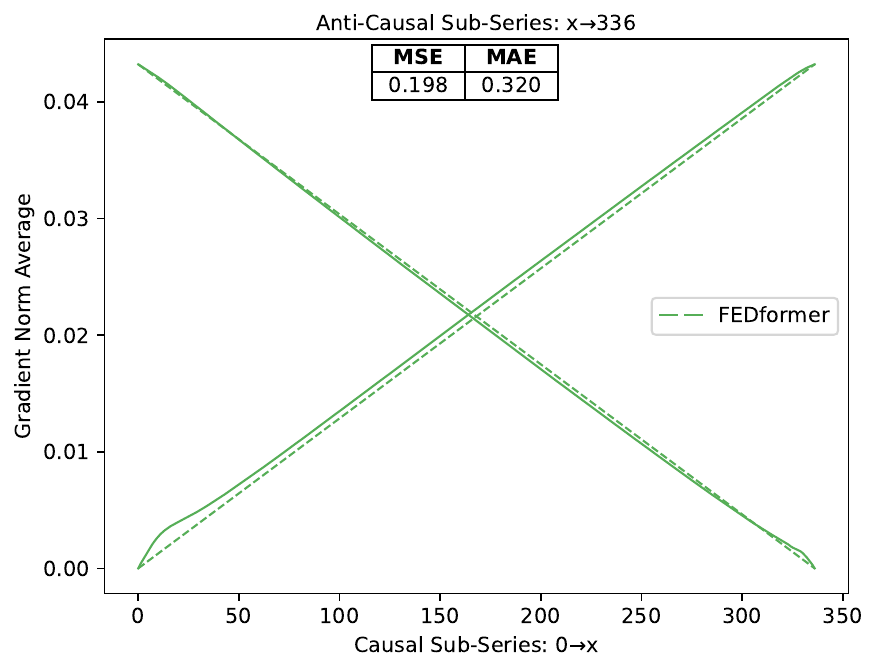}
        };\end{tikzpicture} & \begin{tikzpicture} \node[inner sep=0pt] {
            \includegraphics[width=2.23cm,height=2.1675cm]{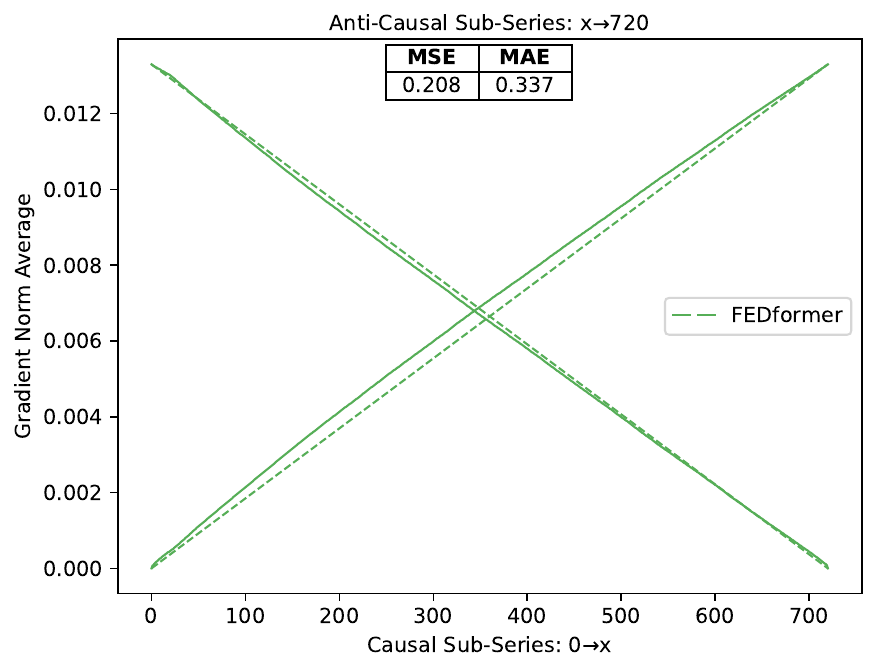}
        };\end{tikzpicture} & \begin{tikzpicture} \node[inner sep=0pt] {
            \includegraphics[width=2.23cm,height=2.1675cm]{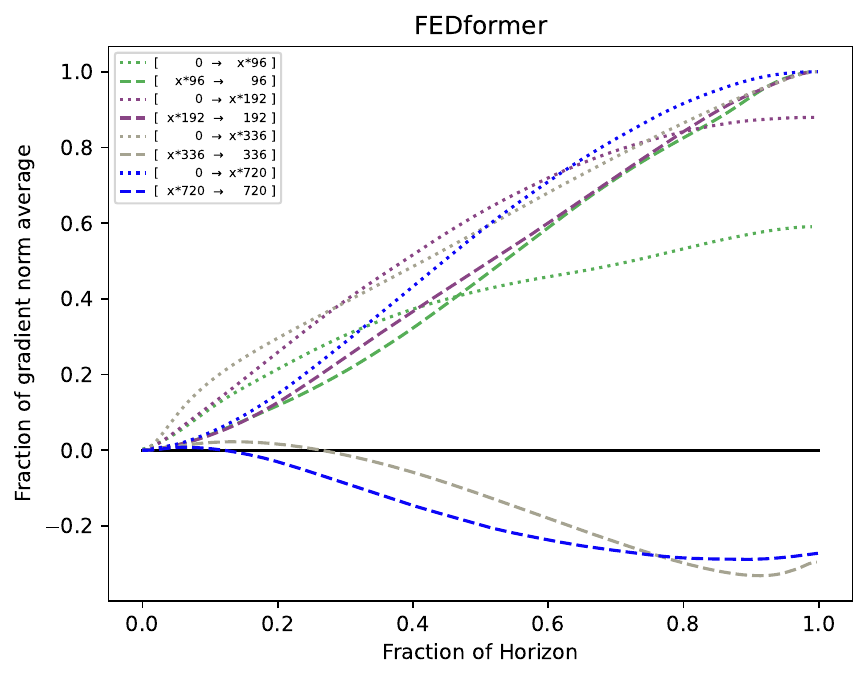}
        };\end{tikzpicture} \\
        \makecell{Pyra-\\former} & \begin{tikzpicture} \node[inner sep=0pt] {
            \includegraphics[width=2.23cm,height=2.1675cm]{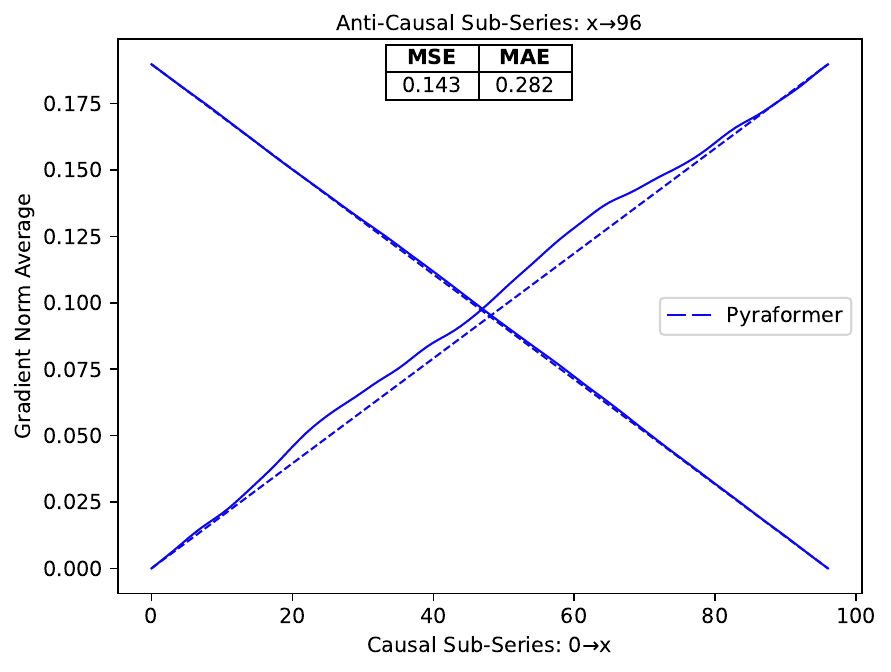}
        };\end{tikzpicture} & \begin{tikzpicture} \node[inner sep=0pt] {
            \includegraphics[width=2.23cm,height=2.1675cm]{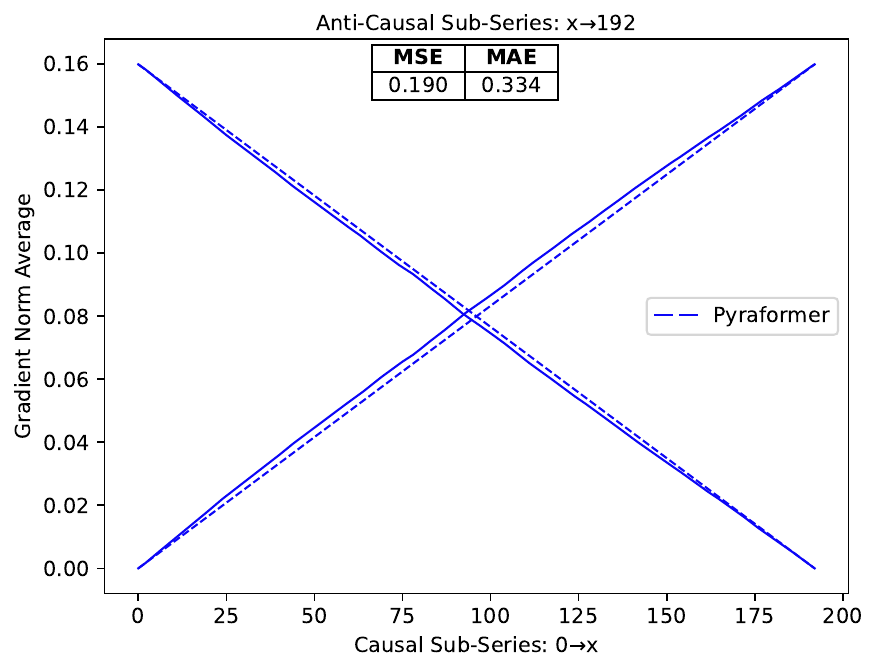}
        };\end{tikzpicture} & \begin{tikzpicture} \node[inner sep=0pt] {
            \includegraphics[width=2.23cm,height=2.1675cm]{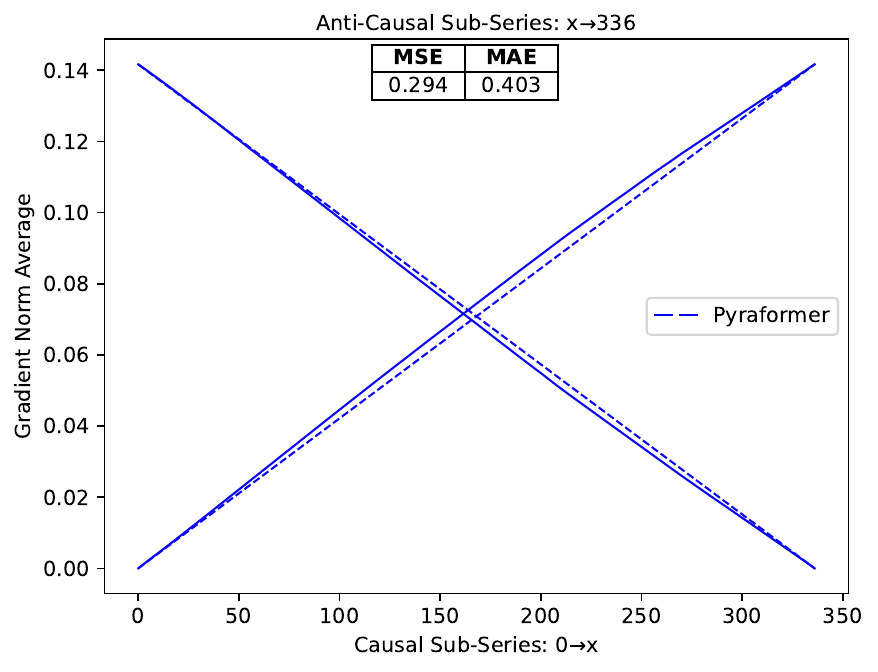}
        };\end{tikzpicture} & \begin{tikzpicture} \node[inner sep=0pt] {
            \includegraphics[width=2.23cm,height=2.1675cm]{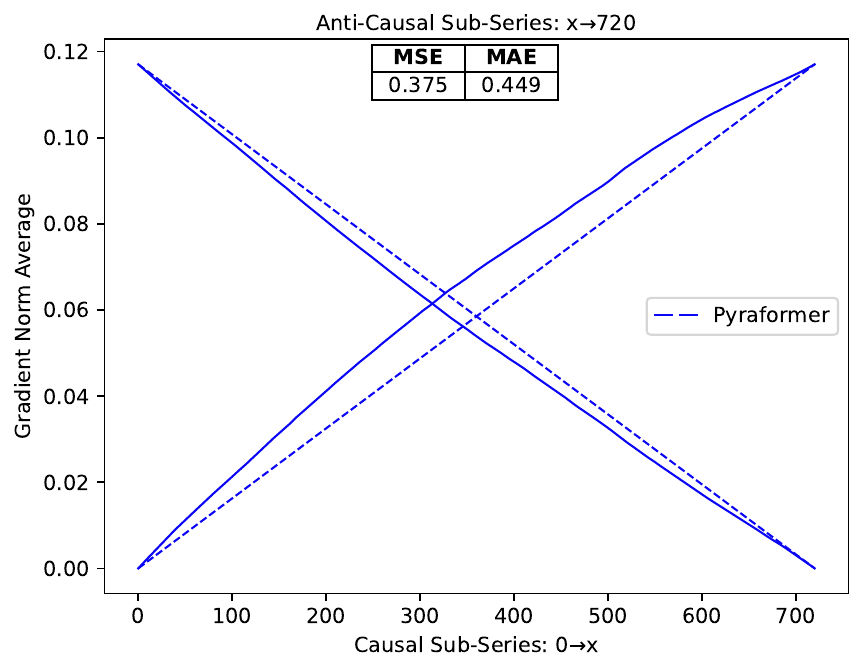}
        };\end{tikzpicture} & \begin{tikzpicture} \node[inner sep=0pt] {
            \includegraphics[width=2.23cm,height=2.1675cm]{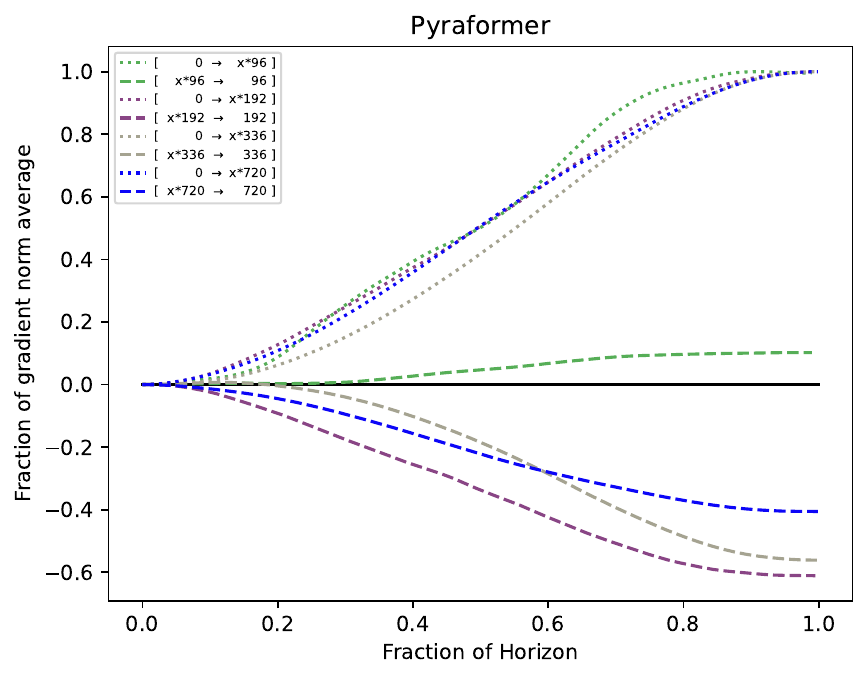}
        };\end{tikzpicture} \\
        \makecell{Space-\\Time} & \begin{tikzpicture} \node[inner sep=0pt] {
            \includegraphics[width=2.23cm,height=2.1675cm]{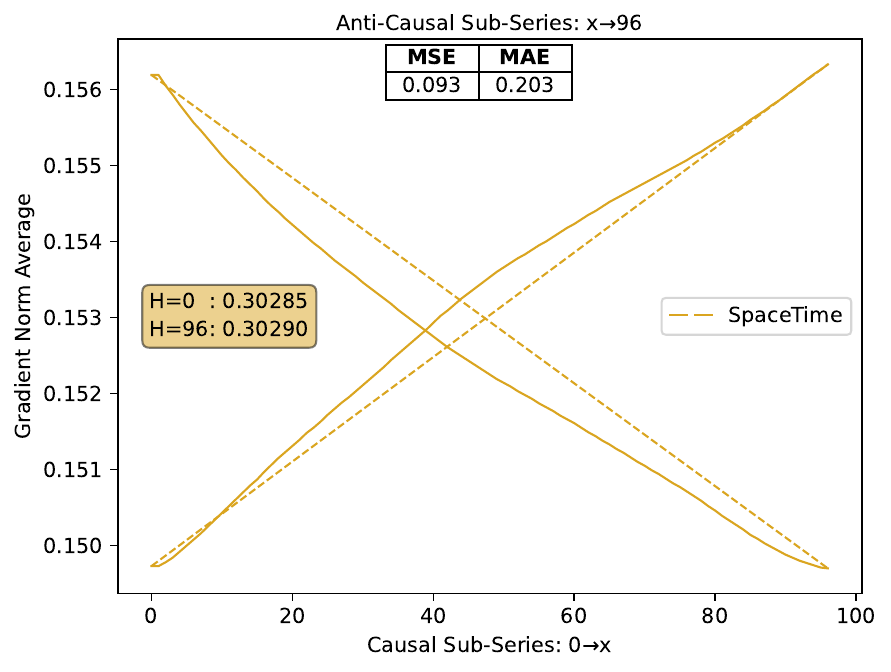}
        };\end{tikzpicture} & \begin{tikzpicture} \node[inner sep=0pt] {
            \includegraphics[width=2.23cm,height=2.1675cm]{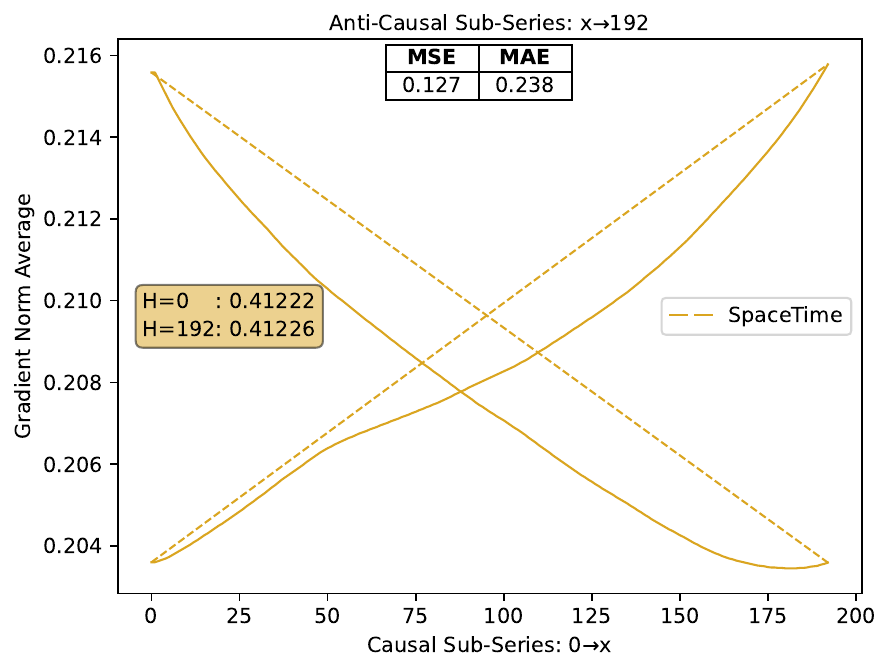}
        };\end{tikzpicture} & \begin{tikzpicture} \node[inner sep=0pt] {
            \includegraphics[width=2.23cm,height=2.1675cm]{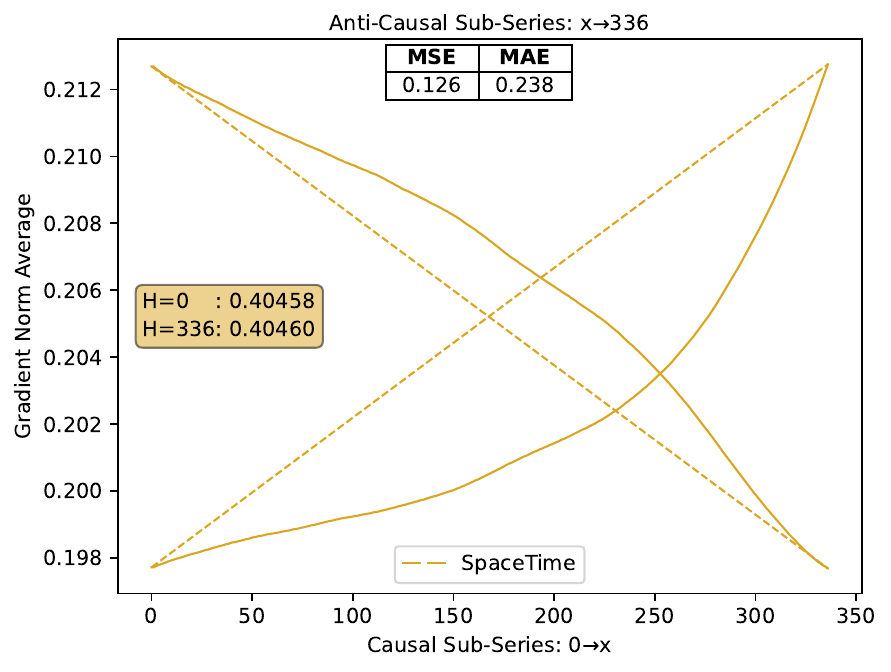}
        };\end{tikzpicture} & \begin{tikzpicture} \node[inner sep=0pt] {
            \includegraphics[width=2.23cm,height=2.1675cm]{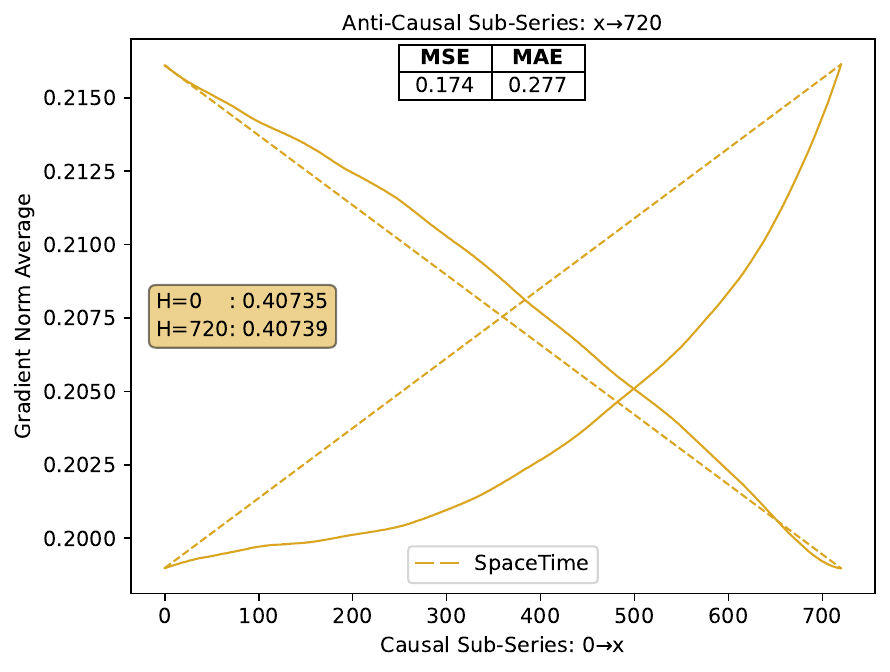}
        };\end{tikzpicture} & \begin{tikzpicture} \node[inner sep=0pt] {
            \includegraphics[width=2.23cm,height=2.1675cm]{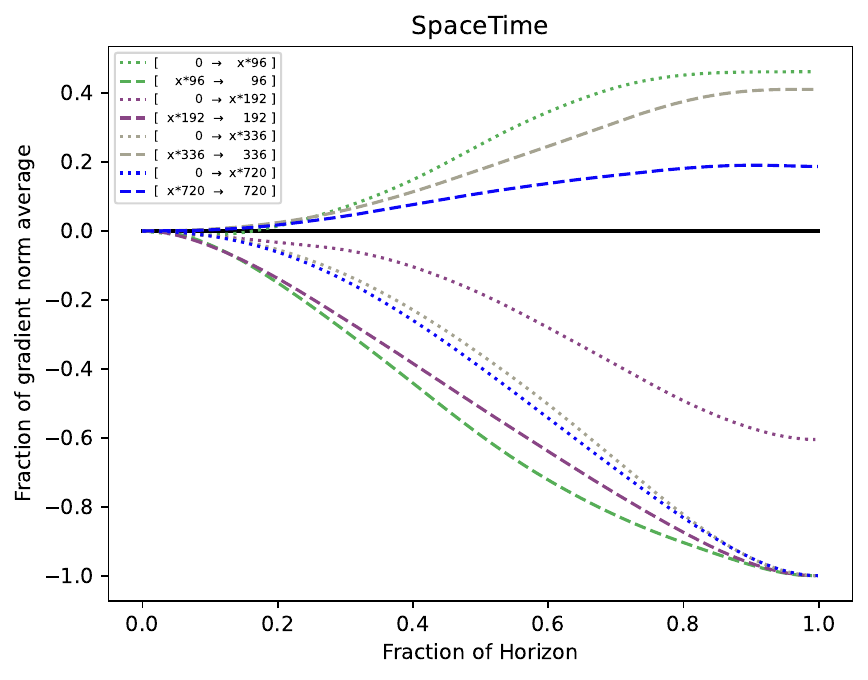}
        };\end{tikzpicture} \\
        \makecell{Multi-\\Reso-\\lution\\DDPM} & \begin{tikzpicture} \node[inner sep=0pt] {
            \includegraphics[width=2.23cm,height=2.1675cm]{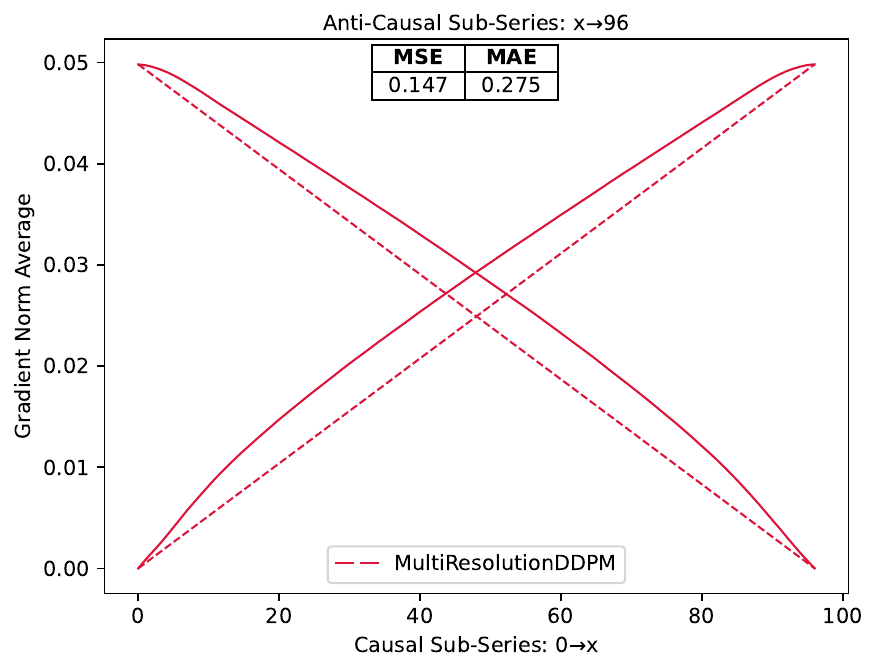}
        };\end{tikzpicture} & \begin{tikzpicture} \node[inner sep=0pt] {
            \includegraphics[width=2.23cm,height=2.1675cm]{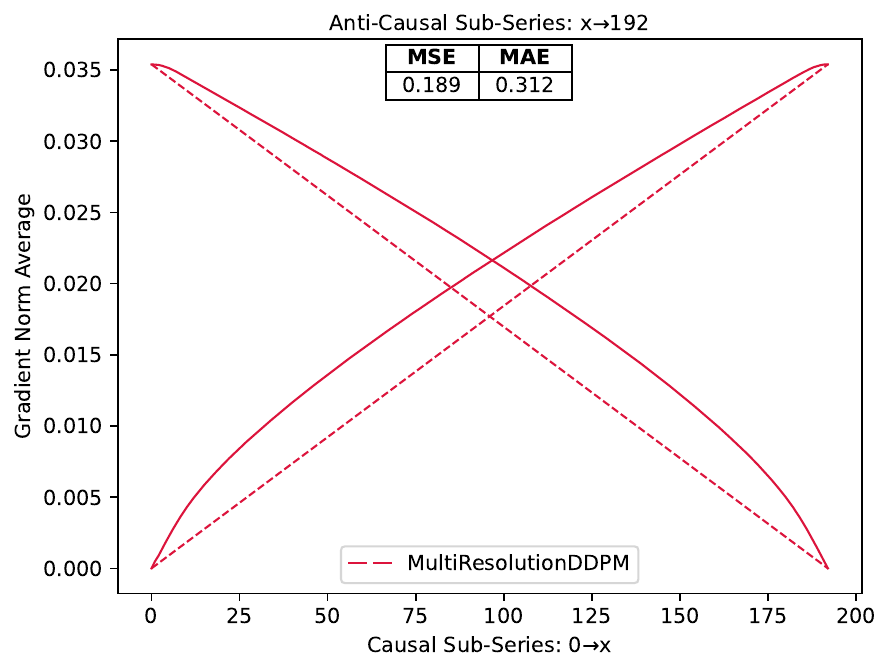}
        };\end{tikzpicture} & \begin{tikzpicture} \node[inner sep=0pt] {
            \includegraphics[width=2.23cm,height=2.1675cm]{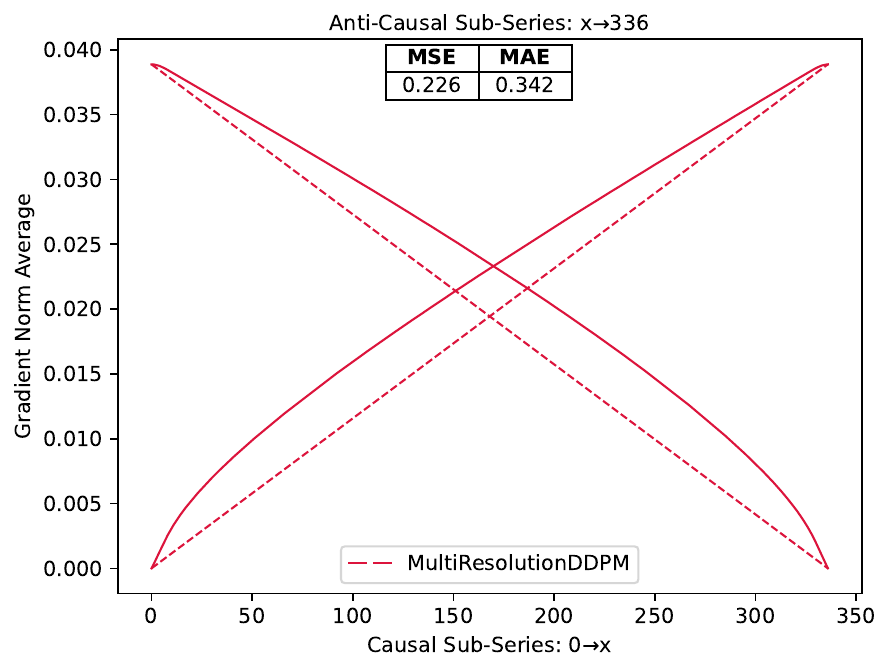}
        };\end{tikzpicture} & \begin{tikzpicture} \node[inner sep=0pt] {
            \includegraphics[width=2.23cm,height=2.1675cm]{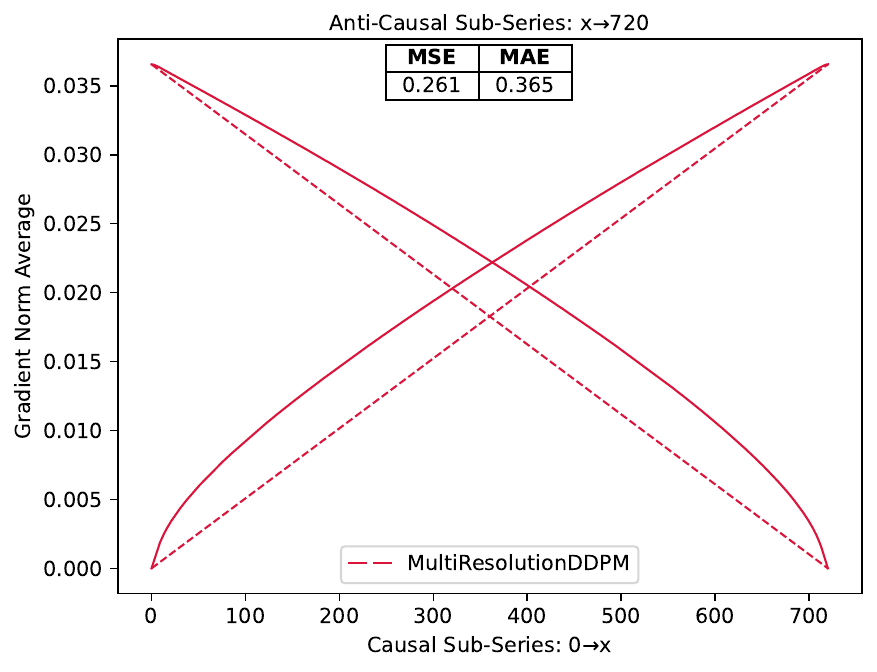}
        };\end{tikzpicture} & \begin{tikzpicture} \node[inner sep=0pt] {
            \includegraphics[width=2.23cm,height=2.1675cm]{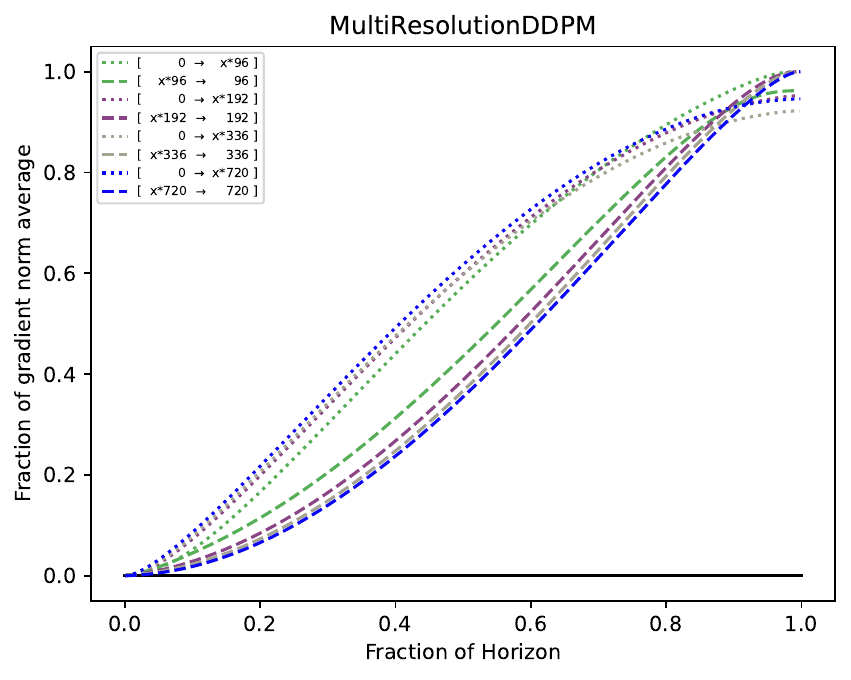}
        };\end{tikzpicture} \\
        \bottomrule
    \end{tabular} 
    \caption{Causal and Anti-causal modes in HAMs and interpolated area plots are plotted for models of horizons $H=\{96,192,336,720\}$.}
    \label{table8}
\end{table}

With \emph{SpaceTime} models, the addition of zeroes to the loss function leads to at least one power of the forecasts' state space weights being 0 in all the cases and reduces magnitudes by 2$\times$. When there are no gradients backpropagated through the forecast branch of the computation graph, the auto-regressive input and layer output based losses make the gradient activations very close to when the entire horizon is forecast, highlighting that HAM decreases magnitudes only in losing timesteps. The $H=0$ and $H=H$ norm average values in the plots are removed and highlighted in the plots to keep them in range. Its causal mode curves transition from almost linear in $H=96$ to exponential trends in its $H=\{192,336,720\}$ models and increase more in the longer regions of the horizon consequently. In becoming exponential causally, its $H=192$ model has the least proportionate anti-causal curve. Its exponential decrease causally enables increase in magnitudes anti-causally that decrease with increase in horizon size. The use of Gaussian noise in \emph{Multi-Resolution DDPM}'s diffusion makes all smaller subseries' gradient norms characteristically larger than proportional in magnitudes with very little variation across horizon sizes and suggest that it could be a property of diffusion. Empirically, the use of magnitudes'-based selections of self-attention parameters using noise representations across layer-wise diffusion in MoDE \cite{reuss2024efficient}, through its gradient updates, found $\sim 40\%$ reduction of parameters and SOTA performances across many long horizon robotics tasks.

\subsection{Difference Plots across Model Families}

\begin{figure}[t]
    \centering
    \subfloat[\centering H=96]{\includegraphics[width=0.495\linewidth,height=2.375cm]{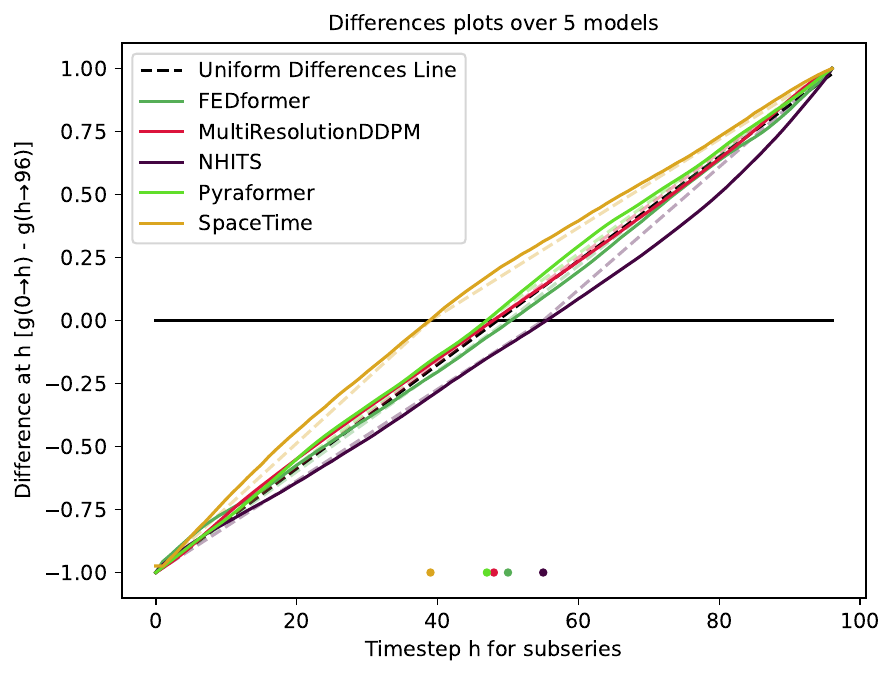}} \hfill
    \subfloat[\centering H=192]{\includegraphics[width=0.495\linewidth,height=2.375cm]{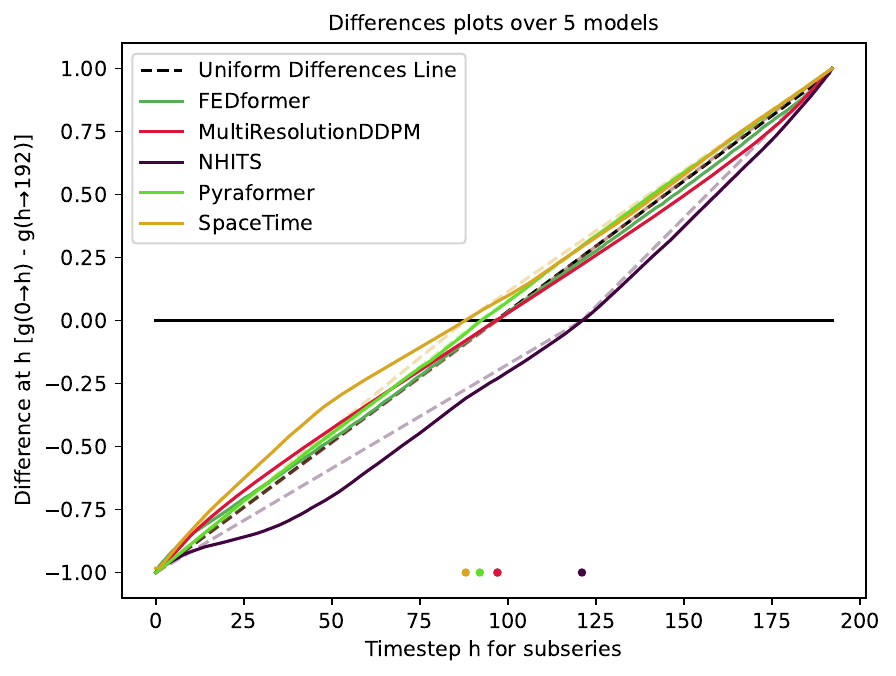}} \\
    \subfloat[\centering H=336]{\includegraphics[width=0.495\linewidth,height=2.375cm]{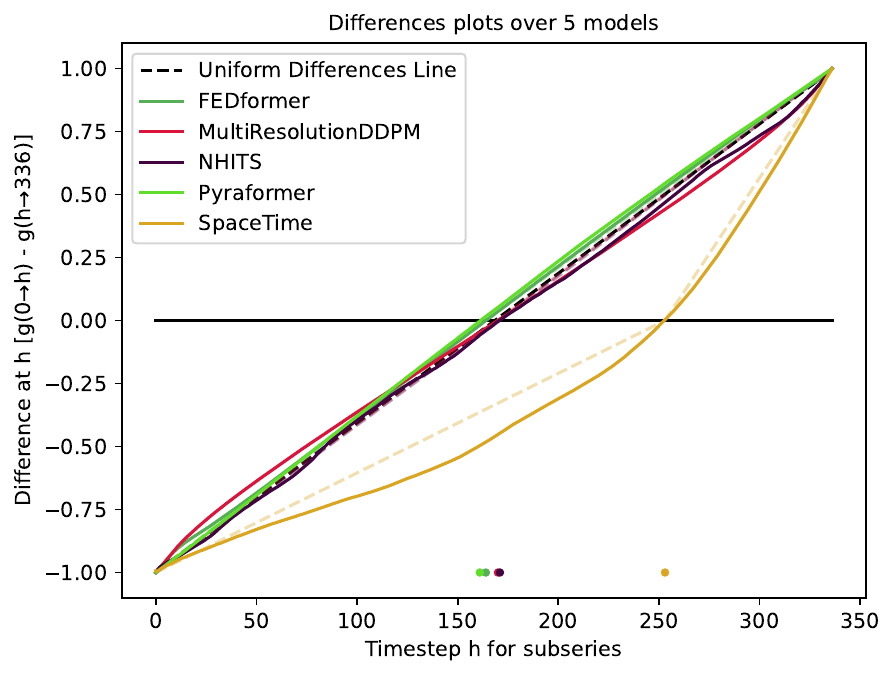}} \hfill
    \subfloat[\centering H=720]{\includegraphics[width=0.495\linewidth,height=2.375cm]{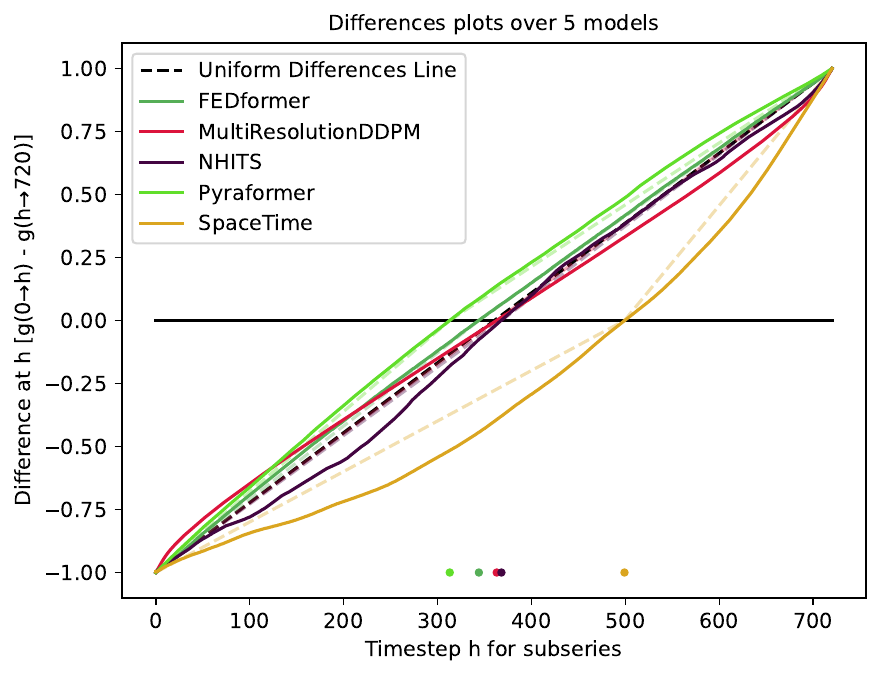}} 
    \caption{Difference plots between gradient norm averages of first $h$ and last $H-h$ timesteps in a horizon of size $H$ with gradient equivariant points below them.}
    \label{diffs}
\end{figure}

\emph{Pyraformer}'s difference curves address long horizon forecasting as an activity magnitude-based emphasis of shorter horizons and correspond to its largest errors across all models. \emph{Multi-Resolution DDPM}'s curves decrease more after its gradient equivariant points in an increasing order of horizon size with curves before them remaining more than proportionate. Its equivariant points are all closest to the $\frac{H}{2}$ timesteps highlighting that they are able to use noise distributions to address long horizon forecasting in terms of the norm average magnitudes. \emph{FEDformer}'s difference plots decrease in magnitudes after equivariant points between $H={96,192}$ and $H={336,720}$ models indicating the robustness of its self-attention in adapting to data distributions and decreases corresponding to longer regions in the horizon as necessary. \emph{SpaceTime}'s difference curves emphasize increases in shorter regions of the horizon decreasing in $H=\{96,192\}$ models and along with its gradient equivariant points being farthest away in the horizons, use its weights' power distributions to exponentially highlight longer regions of the horizon using relative activity magnitudes. \emph{NHITS}' difference plots resemble \emph{SpaceTime}'s in its $H={96,192}$ models and show reductions in its anti-causal mode's longer subseries more than its causal mode's when horizon size increases from 336 to 720. The difference plot decreases in magnitude as the causal mode includes the last 18\% and 14\% of the timesteps in $H=336$ and $H=720$ models respectively, favouring longer horizons lesser as horizon sizes increase. 

\bibliographystyle{splncs04}
\bibliography{ref}

\end{document}